\documentclass[10pt,twocolumn,letterpaper]{article}

\usepackage{iccv}
\usepackage{times}
\usepackage{epsfig}
\usepackage{graphicx}
\usepackage{subfig}
\usepackage{graphicx}
\usepackage{appendix}
\usepackage[accsupp]{axessibility}  
\usepackage{amsmath}
\usepackage{amssymb}
\usepackage{booktabs}
\usepackage{soul}
\usepackage{color}
\usepackage{colortbl}
\definecolor{Gray}{gray}{0.9}
\definecolor{baselinecolor}{gray}{0.9}
\definecolor{yellow}{RGB}{255,239,213}

\definecolor{blue}{RGB}{176,226,255}
\newcommand{\tablestyle}[2]{\setlength{\tabcolsep}{#1}\renewcommand{\arraystretch}{#2}\centering\footnotesize}
\usepackage{threeparttable}
\usepackage{pifont} 
\usepackage{multirow}
\usepackage{overpic}
\usepackage{textpos}
\usepackage{array}
\usepackage{makecell}
\usepackage{array}
\newcolumntype{C}[1]{>{\centering}p{#1}}
\usepackage[pagebackref=true,breaklinks=true,letterpaper=true,colorlinks,bookmarks=false]{hyperref}

\usepackage[capitalize]{cleveref}
\crefname{section}{Sec.}{Secs.}
\Crefname{section}{Section}{Sections}
\Crefname{table}{Table}{Tables}
\crefname{table}{Tab.}{Tabs.}

\iccvfinalcopy 

\def\iccvPaperID{9530} 

\ificcvfinal\fi

\begin{document}

\title{ASAG: Building Strong One-Decoder-Layer Sparse Detectors\\ via Adaptive Sparse Anchor Generation}


\author{Shenghao Fu$^{1,3,4}$, Junkai Yan$^{1,4}$, Yipeng Gao$^{1,4}$, Xiaohua Xie$^{1,3,4*}$, Wei-Shi Zheng$^{1,2,3,4}$\thanks{~denotes the corresponding authors.} \\
\normalsize $^1$School of Computer Science and Engineering, Sun Yat-sen University, China, 
\normalsize $^2$Pengcheng Lab, China, \\ 
\normalsize $^3$Guangdong Province Key Laboratory of Information Security Technology, China, \\ 
\normalsize $^4$Key Laboratory of Machine Intelligence and Advanced Computing, Ministry of Education, China\\
{\tt\small \{fushh7, yanjk3, gaoyp23\}@mail2.sysu.edu.cn}, {\tt\small xiexiaoh6@mail.sysu.edu.cn}, {\tt\small wszheng@ieee.org}}

\maketitle
\ificcvfinal\thispagestyle{empty}\fi

\begin{abstract}
   Recent sparse detectors with multiple, \eg six, decoder layers achieve promising performance but much inference time due to complex heads. Previous works have explored using dense priors as initialization and built one-decoder-layer detectors. Although they gain remarkable acceleration, their performance still lags behind their six-decoder-layer counterparts by a large margin. In this work, we aim to bridge this performance gap while retaining fast speed. We find that the architecture discrepancy between dense and sparse detectors leads to feature conflict, hampering the performance of one-decoder-layer detectors. Thus we propose Adaptive Sparse Anchor Generator (ASAG) which predicts dynamic anchors on patches rather than grids in a sparse way so that it alleviates the feature conflict problem. For each image, ASAG dynamically selects which feature maps and which locations to predict, forming a fully adaptive way to generate image-specific anchors. Further, a simple and effective Query Weighting method eases the training instability from adaptiveness. Extensive experiments show that our method outperforms dense-initialized ones and achieves a better speed-accuracy trade-off. The code is available at \url{https://github.com/iSEE-Laboratory/ASAG}.
\end{abstract}

\section{Introduction}
\begin{figure}[t!]
  \centering
  \includegraphics[width=0.99\linewidth]{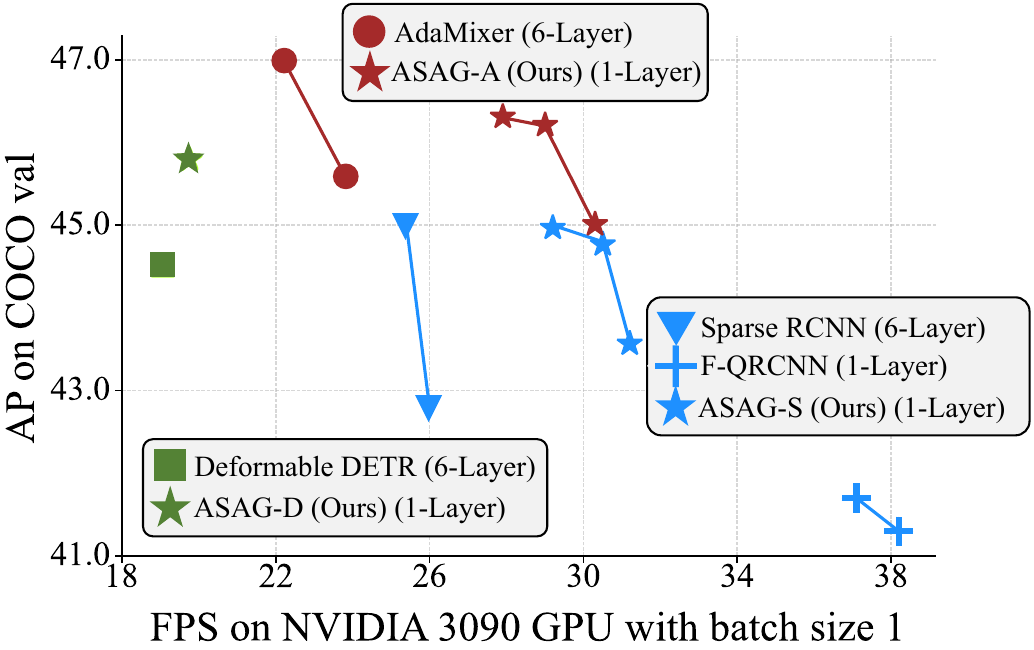}
   \caption{Comparing one-decoder-layer detectors with six-decoder-layer counterparts on FPS and AP with various decoder types. ASAGs achieve a better speed-accuracy trade-off. Best viewed in color.}
   \label{fig:fps_ap}
\end{figure}

Object detection is a fundamental and challenging computer vision task.
Different from traditional CNN-based dense object detectors~\cite{ren2015faster, he2017mask, lin2017feature,lin2017focal, tian2019fcos, atss} using sliding-window paradigm, query-based sparse detectors~\cite{detr, deformabledetr, sun2021sparse, adamixer} use hundreds of object queries to search through the whole image, each representing an object or background. They get rid of some traditional hand-crafted components and procedures, \eg, anchors and Non-Maximum Suppression (NMS), which greatly simplifies the detection pipeline and makes the detector fully end-to-end trainable. With the help of powerful transformer encoder-decoder architecture, sparse detectors show promising performance. 

However, sparse detectors cost more inference time since they need $n$ decoder layers (typically $n=6$) to progressively refine bounding boxes, leading to a more complex head. Inference with only one decoder layer achieves a much faster speed, such as AdaMixer~\cite{adamixer} (43 V.S. 22 FPS) and Sparse RCNN~\cite{sun2021sparse} (43 V.S. 25 FPS). The more complex the decoder is, the larger the FPS gap is. Unfortunately, simply discarding the extra decoder layers results in a significant performance drop.

Recently, several methods~\cite{efficient_detr, zhang2022featurized} have attempted to bridge the performance gap between one- and six-decoder-layer detectors. Efficient DETR~\cite{efficient_detr} finds that image-agnostic box and content queries should be blamed for a severe drop in performance when using one decoder layer. Thus, both methods utilize an extra dense box prediction step before the decoder to provide proper query initialization. However, they fail to achieve comparable results to detectors with six decoder layers, for example Featurized Query RCNN~\cite{zhang2022featurized} is lower than Sparse RCNN~\cite{sun2021sparse} by 1.5AP with 100 queries. We find that the features corresponding to dense and sparse detectors are significantly different. Such feature conflict hampers the performance of one-decoder-layer detectors, demonstrated in \Cref{sec:analysis}. Thus, although these methods achieve remarkable acceleration, it still has much room to narrow the performance gap.

In this work, we aim to build a fully sparse one-decoder-layer detector, narrowing the performance gap between one- and six-decoder-layer detectors and retaining the fast speed. The key difference between our method and other dense initialization methods is that we use patches as the basic prediction units, which can be the whole or part of an image. Sparsely predicting on patches alleviate the feature discrepancy caused by predicting on grids and enjoys global receptive fields. Further, we propose to loose the constraint that each image should use a fixed number of queries to detect objects so that more complex images detect objects with more queries and vice versa. Based on these two preconditions, we propose initializing image-specific queries using Adaptive Sparse Anchor Generator~(ASAG), which is fully adaptive to each image in both anchors' locations and numbers. We further design Adaptive Probing to adaptively crop patches on possible locations on different feature map levels. It runs in a top-down and coarse-to-fine way and greatly enhances the ability to detect small objects. Finally, an effective Query Weighting method is proposed to handle the instability coming from adaptiveness.

We conduct extensive experiments on the COCO~\cite{coco} dataset with various decoder types. As shown in \Cref{fig:fps_ap}, our model ASAG-S outperforms dense-initialized Query RCNN~\cite{zhang2022featurized} by 2.6 AP with fewer FLOPs and the same decoder. We also retain the fast speed of one-decoder-layer detectors, thus achieving a better speed-accuracy trade-off.

\section{Related Works}

\noindent{\textbf{Improving Sparse Detectors.}} Recently, DETR~\cite{detr} views object detection as a set-prediction problem and achieves promising performance. But it still has some apparent disadvantages and lacks interpretability. Many following works have solved the problems like slow convergence and relatively low performance on small objects by utilizing multi-scale feature pyramids~\cite{deformabledetr, adamixer, sam_detr++}, introducing more spatial priors~\cite{deformabledetr, smca_detr, conditional_detr, anchor_detr, dab_detr, adamixer}, stabilizing bipartite matching~\cite{dn_detr, zhang2022dino}, aligning feature space~\cite{sam_detr, sam_detr++}, increasing positive samples~\cite{jia2022detrs, group_detr, h_detr, co_detr, nms_strikes_back}, using knowledge distillation~\cite{teach_detr, detrdistill, d3etr}, initializing queries~\cite{deformabledetr, zhang2022dino, hong2022dynamic}, \etc. 
Although achieving outstanding performance and fast convergence speed, transformer-based sparse detectors still require more inference time than CNN-based dense detectors thus limiting their practical applications.

\vspace{0.5em}\noindent{\textbf{Accelerating Sparse Detectors.}} As a well-known experience, most computation for a detector lies in the backbone due to high-resolution feature maps and dense computation. However, as sparse detectors have more complicated operators in neck and head, \eg, attention and grid sampling, the FLOPs cannot directly reflect the FPS. For example, the decoder of AdaMixer~\cite{adamixer} takes 31\% of the total FLOPs but nearly half of the FPS. Different from works~\cite{act, pnp_detr, sparse_detr} focusing on reducing computation in neck by approximating self-attention in the encoder, we aim to simplify the decoder. Recently, Li \etal~\cite{dpp} find that some unimportant queries are not worth being computed equally. However, decreasing the number of queries brings minor acceleration as Sparse RCNN~\cite{sun2021sparse} increasing the number of queries from 100 to 300 only uses an extra 1 FPS. Thus decreasing the number of decoder stages is a more promising way to speed up sparse detectors.


\begin{table}[t]
  \centering
  \resizebox{\linewidth}{!}{
      \begin{tabular}{c|l|c|cccc}
        \hline
        NO. & Detector & Init. & AP & AP$_{s}$ & AP$_m$ & AP$_l$ \\
        \hline
        (a) & Deformable DETR+~\cite{deformabledetr} & \multirow{2}{*}{learned} & 46.2 & 28.3 & 49.2 & 61.5 \\
        (b) & Deformable DETR+\dag &  & 37.9 & 23.1 & 41.9 & 49.1 \\
        \hline
        (c) & Deformable DETR++~\cite{deformabledetr} & \multirow{2}{*}{dense} & 46.9 & 29.4 & 50.1 & 61.6 \\
        (d) & Deformable DETR++\dag &  & 33.7 & 23.1 & 38.4 & 41.1 \\
        \hline
      \end{tabular}
    }
  \vspace{-0.6em}
  \caption{Effect of dense query initialization on one- and six-decoder-layer detectors. The first decoder layer with dense initialization even underperforms the one without image-specific initialization. \dag: Inference with the first layer.}
  \label{tab:ana_table1}
\end{table}

\section{Why Should We Use Sparse Initialization for One-Decoder-Layer Detectors?}
\label{sec:analysis}

\begin{figure*}[t!]
  \centering
  \includegraphics[width=0.9\linewidth]{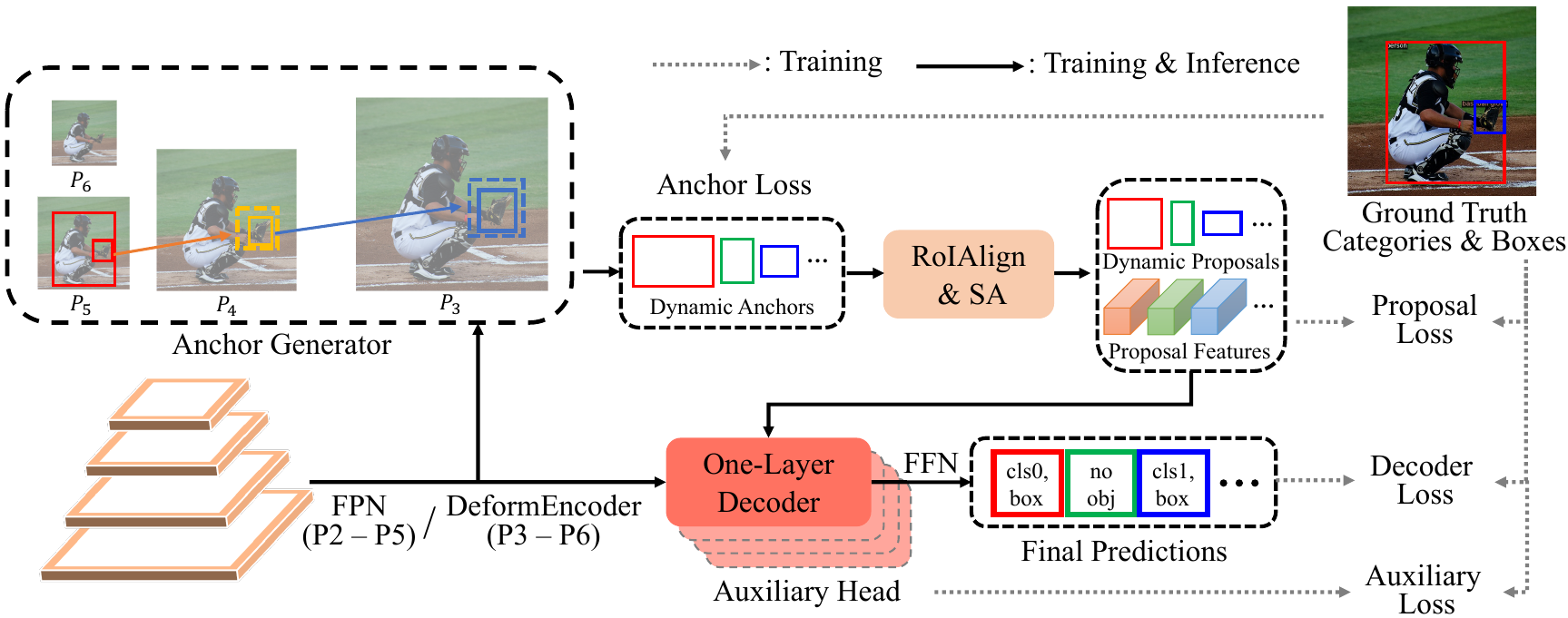}
  \vspace{-0.4em}
   \caption{Overview. Our model first uses Anchor Generator to predict dynamic anchors. Then, query features are extracted by RoIAlign~\cite{he2017mask} and refined by a Self-Attention layer (SA). One decoder layer of any kind is used for final prediction. The neck is changed following the decoder. We additionally add three auxiliary heads using the same proposals to provide more supervision signals, which are discarded during inference. Each component is supervised under one-to-one matching losses.}
   \label{fig:overview}
\end{figure*}

\begin{figure}[t!]
  \centering
  \includegraphics[width=0.99\linewidth]{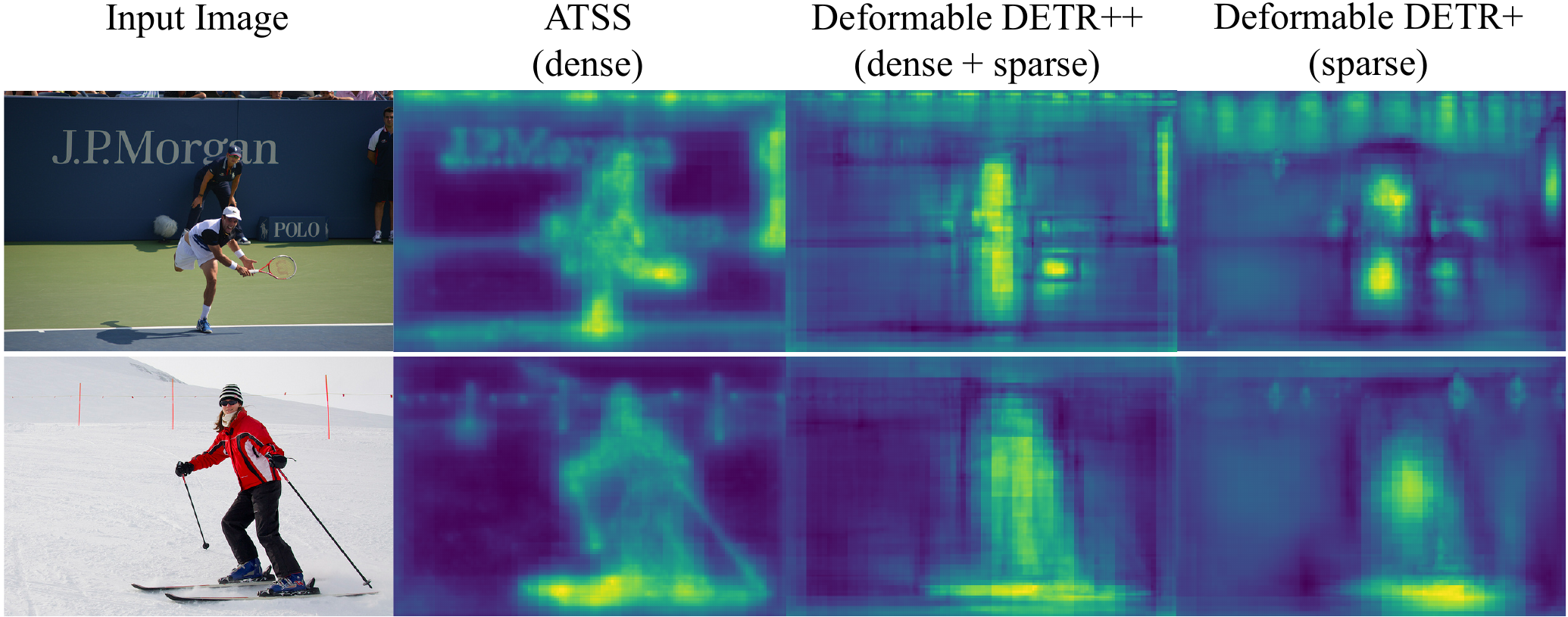}
  \vspace{-0.5em}
   \caption{Comparison on feature maps (discriminability scores~\cite{co_detr}) of dense and sparse detectors. There exists a clear inconsistency between dense and sparse features.}
   \vspace{-0.5em}
   \label{fig:analysis1}
\end{figure}

Some previous works~\cite{deformabledetr, zhang2022dino} have found that utilizing a dense box prediction step as a query initialization can be helpful to six-decoder-layer detectors. For example, Deformable DETR++~\cite{deformabledetr} outperforms Deformable DETR+~\cite{deformabledetr} by 0.7 AP as shown in \Cref{tab:ana_table1}. However, we surprisingly find that the first decoder layer with better initialization even performs worse than the one with image-agnostic initialization, as shown by (b) and (d) in \Cref{tab:ana_table1}. This phenomenon shows that the benefit from the dense box prediction step for six-decoder-layer detectors does not lie in better initialization but in more supervision signals to the encoder, as many works~\cite{group_detr, h_detr, co_detr, nms_strikes_back} find that one-to-one matching is not sufficient for feature learning. 

To better illustrate the phenomenon above, we visualized discriminability scores in the encoder in \Cref{fig:analysis1} following~\cite{co_detr}, which are $l^2$-norm of the corresponding grid features. Objects with higher discriminability scores can be better detected. 
Since dense detectors predict objects based on grid features while sparse detectors detect objects using queries, the architecture discrepancy makes the needed features totally different. As shown in \Cref{fig:analysis1}, sparse detectors pay more attention to background information than dense detectors~\cite{detrdistill}. Moreover, dense detectors prefer to activate the whole object uniformly since each grid has equal chances to predict the object, while sparse detectors tend to highlight some discriminative parts. The score maps of sparse detectors with dense initialization fall between dense and sparse detectors. Due to the powerful decoders with six layers, such detectors with dense initialization can tolerate the discrepancy of features and benefit from more positive signals. However, one-decoder-layer detectors with limited representation ability suffer from conflicting features. We hypothesize that this is why one-decoder-layer detectors with well-initialized queries still lag behind their six-decoder-layer counterparts. This phenomenon also demonstrates that there is redundancy in six-decoder-layer detectors, which can be reduced for acceleration. Thus, we provide a sparse way to further narrow the performance gap between one- and six-decoder-layer detectors and retain the fast speed of one-decoder-layer detectors.

\section{Adaptive Sparse Anchor Generation for Query Initialization}
\label{sec:method}

\subsection{Overview}


In this section, we introduce our Adaptive Sparse Anchor Generator (ASAG for short), which initializes queries sparsely and is more suitable for sparse decoders while retaining fast speed. As shown in \Cref{fig:overview}, ASAG generates image-specific anchors adaptively from the aspect of locations and numbers, without using predefined spatial priors. Unlike complex decoders, our ASAG is lightweight, using only 0.06G FLOPs. With dynamic anchors, we use RoIAlign~\cite{he2017mask} to generate content queries since initializing both box and content queries is vital for one-decoder-layer detectors~\cite{efficient_detr}. Further, an extra Self-Attention layer is utilized to model relationships between objects and reduce redundancy for NMS-free, similar to Featurized QRCNN~\cite{zhang2022featurized}. Lastly, a one-layer decoder of any kind~\cite{adamixer, sun2021sparse, deformabledetr} is used for final refinements. 

Recent works~\cite{jia2022detrs, group_detr, h_detr, co_detr, nms_strikes_back} show that sparse detectors benefit from sufficient supervision signals. Considering that dense methods provide supervision to each grid and other six-decoder-layer detectors have more auxiliary losses, we additionally add three auxiliary parallel one-layer decoders for a fair comparison. Thus, our models are also supervised by six one-to-one matching losses. Different from Group DETR~\cite{group_detr} that uses different groups of queries and the shared decoder by each group, we use different decoders with the same proposals.

\begin{figure}[t!]
  \centering
  \includegraphics[width=0.99\linewidth]{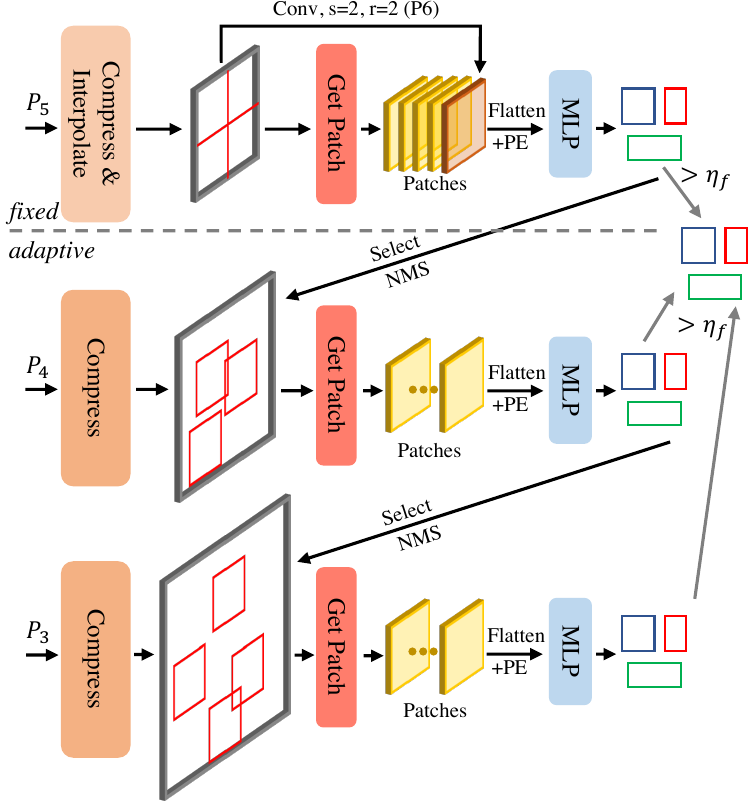}
  \vspace{-0.3em}
   \caption{Adaptive Sparse Anchor Generator (ASAG). ASAG starts predicting dynamic anchors from fixed feature maps and then adaptively explores larger feature maps using Adaptive Probing, which runs top-down and coarse-to-fine. Learnable position embeddings (PE) are added to patches after flattening to keep spatial structures.}
   \vspace{-0.4em}
   \label{fig:anchor_generator}
\end{figure}

\subsection{Adaptive Sparse Anchor Generator}
\label{subsec:Adaptive_Sparse_Anchor_Generator}

\noindent{\textbf{Patches are better prediction units.}} 
In this work, we use \textbf{patches} as the basic prediction units, from which possible anchors will be predicted. The patch is the whole or part of an image, which may contain lots of objects, and is much larger than grids or regions of interest. Thus, predicting from patches alleviates the feature discrepancy caused by predicting from grids and enjoys global receptive fields.

\vspace{0.5em}\noindent{\textbf{Inference with the fixed number of feature maps.}}
We start demonstrating our method from using a fixed number of feature maps to predict all objects, \ie $P_5$\footnote{$P_l$ denotes the feature map downsampled by a factor of $2^{l}$.} and $P_6$. As shown in the upper part of \Cref{fig:anchor_generator}, taking the $P_5$ feature map as input, we first compress the feature map to a few channels along the channel dimension to save computation and parameters. 
To handle images with different sizes, we interpolate $P_5$ to a fixed size and then evenly split it into four patches from top-left to bottom-right to narrow the predictor's search space. If we directly view the whole $P_5$ feature map as a single patch, the model tends to overlook some small objects. Considering that some large objects may appear across the patches, $P_6$ feature map is used to handle the problem, which is downsampled from $P_5$ after interpolating by a factor of 2, and viewed as a single patch (the brown patch). We use an MLP as the predictor to predict anchors of a fixed number on each patch simultaneously, each described by four coordinates and a location score. The location score can be seen as the class probability so that it is class-agnostic and is supervised by IoU~\cite{gfl} as a soft label.

\vspace{0.5em}\noindent{\textbf{Inference with a dynamic number of feature maps.}} Since $P_5$ is not sufficient to predict small objects, using feature maps with larger sizes can obtain more precise anchors. Motivated by QueryDet~\cite{yang2022querydet}, we propose to sparsely compute on large feature maps using \textbf{Adaptive Probing} to correct the anchors for small objects with low confidence. 

As shown in the lower part of \Cref{fig:anchor_generator}, we select some anchors predicted from the fixed part, whose confidences fall in [$\eta_l$, $\eta_h$] and sizes are smaller than half of the patch size. Anchors with scores lower than $\eta_l$ are seen as noisy anchors, and anchors with scores higher than $\eta_h$ are accurate enough. Thus such anchors are not used in Adaptive Probing. With selected anchors, we crop some patches on the $P_4$ feature map, whose centers are the corresponding anchors' ones. Since patches may overlap with each other, we use NMS with IoU threshold $\eta_{iou}$ to reduce the redundancy. Anchors predicted from the higher resolution feature maps are more precise than the original ones, thus we replace them with the newly generated ones. Such probing iterations continue to perform on the following larger feature maps until the largest one $P_3$. We empirically find that additionally using $P_2$ brings minor improvement but adds more inference time. Once no anchors are selected, the iterations break with an early-stop mechanism. Thus, the probing is adaptive to the number of iterations, the number of patches, and the locations of patches. Finally, all anchors with scores higher than $\eta_f$ and not being selected for Adaptive Probing are gathered. Considering that the model needs to deal with pictures with different difficulties, the number of generated patches and anchors varies accordingly. We pad the output anchors to the max size for parallel processing.

\vspace{0.5em}\noindent{\textbf{Training ASAG.}} We first define three kinds of patches: 1) \emph{generated patch} which is generated from selected anchors, 2) \emph{GT patch} which is gained by grouping ground truth boxes smaller than half of the patch size, and 3) \emph{random patch} which is generated randomly. To ensure that predictors for each pyramid level are fully and equally trained, we define a minimal training patch number $N_{TP}=4$ for each level since the lower level tends to receive fewer supervision signals due to the early-stop mechanism. For feature maps from $P_5$ to $P_3$, we use the generated patch, GT patch, and random patch in turn until the minimum patch number is met. For $P_6$, since we view the whole feature map as a single patch, we flip the patch horizontally and vertically to meet the minimum patch number. Only anchors generated from generated patches with confidence scores higher than $\eta_f$ are gathered and sent to the following model parts. The targets for each patch are objects whose centers lie in the patch. Since the output anchors are unordered, bipartite matching is used to get one-to-one matching, similar to other parts of our model and other sparse detectors, except it is class-agnostic. Further, we use IoUs between ground truth boxes and the matched anchors as the soft labels~\cite{gfl}. 

\vspace{0.5em}\noindent{\textbf{Relationships with related works.}} The Adaptive Probing computes sparsely on large feature maps to save computation, similar to PointRend~\cite{pointrend} and QueryDet~\cite{yang2022querydet}. However, both are dense prediction methods and it is clear where to explore on the larger feature map while we sparsely find the corresponding location. Further, QueryDet predicts objects in a \textbf{divide-and-conquer} way, but the Adaptive Probing is in a \textbf{correct-and-replace} manner thus we enjoy the early-stop mechanism. We can even discard large feature maps manually for efficient inference, as shown in \Cref{tab:compare_1x}.

\subsection{Stablizing Training}
\begin{figure}[t!]
  \centering
  \includegraphics[width=0.99\linewidth]{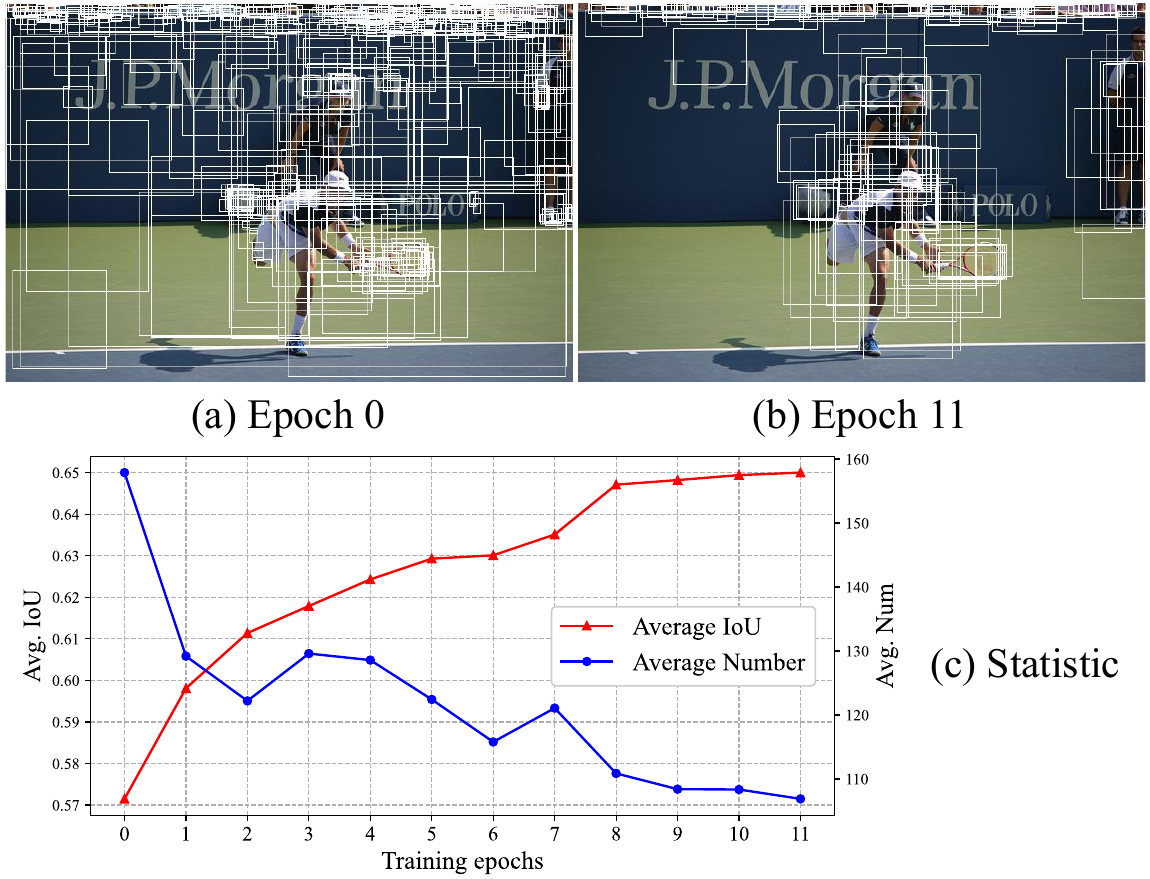}
  \vspace{-0.2em}
   \caption{Visualizing dynamic anchors during the training process. The quality and number of anchors change along the training process making the model hard to optimize.}
   \label{fig:stable}
\end{figure}

Through the novel design above, we gain adaptive sparse anchors and proposals efficiently. However, as shown in \Cref{fig:stable}, we find that our dynamic anchors have two characteristics that are significantly different from the traditional hand-crafted anchors: 1) dynamic anchors may not be as precise as predefined anchors in the early training stage, 2) dynamic anchors change both in quality and numbers along the training process, making the detection head hard to optimize. It is unsuitable for treating two anchors with 0.1 and 0.9 IoU equally since it confuses the detectors about the definition of positive samples. Thus, we propose \textbf{Query Weighting} to ease the training difficulty by giving high-quality anchors with larger weights and vice versa. Soft labels make detectors pay more attention to precise predictions, stabilizing the training when dynamic anchors change, especially in the early training process. This simple weighting mechanism introduces no inference cost.

Motivated by DW~\cite{dw} to give diverse positive and negative loss weights, our weighting functions are as follows:
\begin{align}
&\text{Norm}(x_1, x_2) = \sigma((x_1\times x_2 - 1/3) \times 4.5) \div \sigma(3), \label{eq:norm}\\
&w_{pos} = \text{Norm}(s^{\gamma_1}, IoU^{\gamma_2}), \label{eq:w_pos} \\
&w_{neg} = \text{Norm}(s^{\gamma_1}, P_{neg}(IoU^{\gamma_2})) - \sigma(-1.5), \label{eq:w_neg}
\end{align}
where $s$ and $IoU$ are classification scores $s$ and IoUs, $P_{neg}$ is the same function as in~\cite{dw}, and $\sigma$ denotes sigmoid function. After normalizing, the positive weights are roughly in [0.2, 1] and the negative weights are in [0, 0.8]. Since the sigmoid function is non-linear, it raises the small values while still keeping them within [0, 1]. Even if the matched anchors do not overlap with the targets, we cannot assign the positive weights to zeros as no other anchors will be assigned to the targets in the one-to-one label assignment. And we avoid assigning ones to negative weights of the only matched anchors. The weights only apply to losses not to matching costs. 

Different from label weighting methods~\cite{gfl, tood, dw} in dense detectors, which align the separate regression and classification heads by giving larger weights to more suitable anchors within the candidate bag of each ground truth, Query Weighting is to tolerate dynamic anchors during the training process. Thus our Query Weighting is global-wise while label weighting is instance-wise. We show that six-decoder-layer detectors do not benefit from Query Weighting due to regressing from fixed queries in \Cref{sec:discussion}.

\section{Experiments}
\label{sec:Experiments}

\subsection{Settings}
\noindent{\textbf{Dataset.}} We conduct ASAG experiments on the widely used detection dataset COCO2017~\cite{coco}. We train our model on \texttt{train2017} ($\sim$118k images) and report evaluation metrics on \texttt{val2017} containing 5k images. We find each parallel decoder performs similarly ($\sim$0.2 AP).

\vspace{0.5em}\noindent{\textbf{Configurations.}} To show the generalizability and make a fair comparison, we conduct experiments with well-known decoders, such as AdaMixer~\cite{adamixer}, Sparse RCNN~\cite{sun2021sparse}, Deformable DETR~\cite{deformabledetr}, and our corresponding models are ASAG-A, ASAG-S, and ASAG-D, respectively. Featurized Query RCNN~\cite{zhang2022featurized} and Efficient DETR~\cite{efficient_detr} are dense-initialized one-decoder-layer detectors of Sparse RCNN and Deformable DETR, respectively. We compare with these methods using a similar average number of anchors fairly. 
For the 100 queries setting, we set the range of the number of anchors for each image as [5, 200], $\eta_f$ as 0.1, the number of predicted anchors for each patch on the fixed part and the adaptive part as 50 and 20. The corresponding configurations for the 300 queries setting are [50, 500], 0.05, 150, and 50.

\vspace{0.5em}\noindent{\textbf{Other implementation details.}} We keep the initial learning rate for the backbone and other parts as $2\times 10^{-5}$ and $2\times 10^{-4}$. The batch size is 16. In the standard 1$\times$ schedule, the learning rate drops at epoch 8 and 11 by a factor of 0.1. In the 3$\times$ schedule, multi-scale training is utilized and the shorter side of images ranges from 640 to 800. The learning rate drops at epoch 24 and 33. No query patterns~\cite{anchor_detr} is used. Following other DETR-like models, we use L1 loss, GIoU loss~\cite{giou}, and classification losses with coefficients 5, 2, 2. The Query Weighting applies to both regression and classification losses similar to~\cite{dw}. AdamW~\cite{adamw} with weight decay 0.0001 is used as the optimizer.

As for other default hyper-parameters, the patch size in Anchor Generator is 15, thus the interpolate size for $P_5$ is 30. $\gamma_1$ and $\gamma_2$ in Query Weighting are 0.4 and 0.6. For fast inference, we stop Adaptive Probing with an early stop if the number of selected anchors is less than 3. And we limit the number of patches for each level within 15. 
FPS are tested on a single NVIDIA 3090 GPU with batch size 1.

\subsection{Main Results}
\label{sec:main_results}

\begin{table}[t]
  \centering
  \resizebox{0.99\linewidth}{!}{
      \begin{tabular}{l|c|c|c|m{0.4cm}<{\centering}m{0.4cm}<{\centering}m{0.4cm}<{\centering}m{0.4cm}<{\centering}m{0.4cm}<{\centering}m{0.5cm}<{\centering\arraybackslash}|c}
        \hline
        Detector & FL & \#An & \#L & AP & AP$_{50}$ & AP$_{75}$ & AP$_{s}$ & AP$_m$ & AP$_l$ & FPS \\
        \hline
        \multirow{2}{*}{AdaMixer} &\cellcolor{blue}  - & \cellcolor{blue} 100 & \cellcolor{blue} 6 & \cellcolor{blue} 42.7 & \cellcolor{blue} 61.5 & \cellcolor{blue} 45.9 & \cellcolor{blue} 24.7 & \cellcolor{blue} 45.4 & \cellcolor{blue} 59.2 & \cellcolor{blue} 23.8 \\
        \cline{2-11}
         & \cellcolor{blue} - & \cellcolor{blue} 300 & \cellcolor{blue} 6 & \cellcolor{blue} 44.1 & \cellcolor{blue} 63.4 & \cellcolor{blue} 47.4 & \cellcolor{blue} 27.0 & \cellcolor{blue} 46.9 & \cellcolor{blue} 59.5 & \cellcolor{blue} 22.2 \\
        \hline
        \multirow{6}{*}{ASAG-A} & $P_{3\text{-}6}$ & 107 & 1 & 42.6 & 60.5 & 45.8 & 25.9 & 45.8 & 56.9 & 28.9 \\
         & \cellcolor{yellow} $P_{4\text{-}6}$ & \cellcolor{yellow} 97 & \cellcolor{yellow} 1 & \cellcolor{yellow} 42.1 & \cellcolor{yellow} 59.9 & \cellcolor{yellow} 45.7 & \cellcolor{yellow} 24.8 & \cellcolor{yellow} 46.0 & \cellcolor{yellow} 56.9 & \cellcolor{yellow} 30.3 \\
         & \cellcolor{yellow}  $P_{5\text{-}6}$ & \cellcolor{yellow} 87 & \cellcolor{yellow} 1 & \cellcolor{yellow} 40.3 & \cellcolor{yellow} 57.5 & \cellcolor{yellow} 43.3 & \cellcolor{yellow} 21.2 & \cellcolor{yellow} 44.4 & \cellcolor{yellow} 57.1 & \cellcolor{yellow} 31.3 \\
        \cline{2-11}
         & $P_{3\text{-}6}$ & 329 & 1 & 43.6 & 62.5 & 47.0 & 26.9 & 46.2 & 57.6 & 27.9 \\
         & \cellcolor{yellow} $P_{4\text{-}6}$ & \cellcolor{yellow} 313 & \cellcolor{yellow} 1 & \cellcolor{yellow} 43.4 & \cellcolor{yellow} 62.1 & \cellcolor{yellow} 46.8 & \cellcolor{yellow} 26.4 & \cellcolor{yellow} 46.4 & \cellcolor{yellow} 57.7 & \cellcolor{yellow} 29.0 \\
         & \cellcolor{yellow} $P_{5\text{-}6}$ & \cellcolor{yellow} 280 & \cellcolor{yellow} 1 & \cellcolor{yellow} 41.9 & \cellcolor{yellow} 60.5 & \cellcolor{yellow} 44.9 & \cellcolor{yellow} 23.5 & \cellcolor{yellow} 45.5 & \cellcolor{yellow} 57.9 & \cellcolor{yellow} 30.5 \\
        \hline
      \end{tabular}
    }
    \vspace{-0.4em}
  \caption{Comparison with AdaMixer~\cite{adamixer} using the standard 1$\times$ schedule and R50. FL means feature map levels used in Anchor Generator. \#An and \#L denote the average number of anchors and the number of decoder layers, respectively. Models colored in \colorbox{yellow}{yellow} use efficient inference, and in \colorbox{blue}{blue} use more than one decoder layer. Using only $P_{5\text{-}6}$ means do not use Adaptive Probing totally.}
  \vspace{-0.2em}
  \label{tab:compare_1x}
\end{table}

\noindent{\textbf{Comparison under the standard 1$\times$ schedule.}} We first fairly compare ASAG-A with AdaMixer~\cite{adamixer} with a few epochs. As shown in \Cref{tab:compare_1x}, although dynamic anchors are changing and imprecise in the early time, Query Weighting stabilizes the training and ASAG-A still converges in 12 epochs and achieves comparable results with six-decoder-layer AdaMixer with 1.25$\times$ speed-up. Since Adaptive Probing runs in an iterative and correct-and-replace way, we can stop at a specific level manually for efficient inference without re-training. Note that efficient inference will not hurt the performance of large objects.

\vspace{0.5em}\noindent{\textbf{Comparison with one-decoder-layer detectors.}} Since Anchor Generator alleviates the feature discrepancy caused by predicting from grids, ASAG-D outperforms Efficient DETR~\cite{efficient_detr} by 0.7 AP and ASAG-S outperforms Featurized Query RCNN~\cite{zhang2022featurized} by 2.6 AP, as shown in \Cref{tab:compare_one_layer_3x}, showing the effectiveness of ASAG. Besides, different from computing densely on large feature maps, ASAG sparsely selects patches on different feature maps, saving much computation. Further, since the patch on $P_6$ is the whole image, ASAG enjoys global receptive fields, bringing about much higher AP$_l$ compared with dense(grid)-initialized ones.

\vspace{0.5em}\noindent{\textbf{Comparison with six-decoder-layer detectors with the same decoder type.}}
As shown in \Cref{tab:compare_3x}, while existing one-decoder-layer detectors with dense initialization fall behind their six-decoder-layer counterparts by a large margin, our model greatly narrows the performance gap. In particular, our ASAG-S even outperforms Sparse RCNN~\cite{sun2021sparse} in the 100 queries setting with faster speed and fewer FLOPs. More comparisons with other well-known detectors can be found in supplementary materials.

\begin{table}[t]
  \centering
  \resizebox{\linewidth}{!}{
      \begin{tabular}{ll|c|c|m{0.4cm}<{\centering}m{0.4cm}<{\centering}m{0.4cm}<{\centering}m{0.4cm}<{\centering}m{0.4cm}<{\centering}m{0.5cm}<{\centering}|c|c}
        \toprule
        \multicolumn{2}{l|}{Detector} & \#An & \#L & AP & AP$_{50}$ & AP$_{75}$ & AP$_{s}$ & AP$_m$ & AP$_l$ & F & FPS \\
        \hline
        
        \multicolumn{2}{l|}{F-QRCNN~\cite{zhang2022featurized}} & 100 & 1 & 41.3 & 59.4 & 44.9 & 26.7 & 44.2 & 52.4 & 140 & 38.2 \\
        \multicolumn{2}{l|}{F-QRCNN*~\cite{zhang2022featurized}} & 300 & 1 & 41.7 & 60.2 & 45.4 & 27.7 & 44.3 & 52.0 & 143 & 37.1 \\
        \cline{2-12}

        ASAG-S & $P_{3\text{-}6}$ & 100 & 1 & 43.9 & 62.2 & 47.9 & 27.8 & 46.5 & 57.8 & 130 & 30.1 \\
        \rowcolor{yellow} \cellcolor{white} (Ours) & $P_{4\text{-}6}$ & 89 & 1 & 43.6 & 61.7 & 47.5 & 26.7 & 46.7 & 57.8 & 130 & 31.2 \\
         \rowcolor{yellow} \cellcolor{white} & $P_{5\text{-}6}$ & 76 & 1 & 41.5 & 58.9 & 45.2 & 23.2 & 45.4 & 57.8 & 130 & 33.0 \\

         \cline{2-12}

        ASAG-S & $P_{3\text{-}6}$ & 312 & 1 & 45.0 & 64.1 & 49.1 & 29.5 & 47.4 & 57.8 & 136 & 29.2 \\

        \rowcolor{yellow} \cellcolor{white} (Ours) & $P_{4\text{-}6}$ & 292 & 1 & 44.8 & 63.9 & 48.8 & 28.9 & 47.5 & 57.8 & 136 & 30.5 \\
        \rowcolor{yellow} \cellcolor{white}  & $P_{5\text{-}6}$ & 256 & 1 & 43.2 & 61.9 & 47.1 & 25.8 & 46.7 & 57.8 & 136 & 31.7 \\
        \hline
        \multicolumn{2}{l|}{Effi-DETR~\cite{efficient_detr}} & 300 & 1 & 45.1 &  63.1 & 49.1 & 28.3 & 48.4 & 59.0 & 210 & - \\
        \multicolumn{2}{l|}{ASAG-D (Ours)} & 253 & 1 & 45.8 & 64.1 & 49.4 & 27.3 & 49.6 & 61.0 & 182 & 19.7 \\
        \hline
        \multicolumn{2}{l|}{ASAG-A (Ours)} & 102 & 1 & 45.3 & 63.3 & 48.9 & 27.3 & 48.5 & 59.7 & 131 & 28.9 \\
        \multicolumn{2}{l|}{ASAG-A (Ours)} & 312 & 1 & 46.3 & 65.1 & 50.3 & 29.9 & 49.2 & 59.6 & 139 & 27.9 \\
        \bottomrule
      \end{tabular}
    }
  \vspace{-0.8em}
  \caption{Comparison with other one-decoder-layer detectors with 36 epochs and R50. *: Reimplement by us using official codes. F denotes GFLOPs.}
  \label{tab:compare_one_layer_3x}
\end{table}

\begin{table}[t]
  \centering
  \resizebox{\linewidth}{!}{
      \begin{tabular}{l|c|c|m{0.4cm}<{\centering}m{0.4cm}<{\centering}m{0.4cm}<{\centering}m{0.4cm}<{\centering}m{0.4cm}<{\centering}m{0.5cm}<{\centering}|c|c}
        \toprule
        Detector & \#An & \#L & AP & AP$_{50}$ & AP$_{75}$ & AP$_{s}$ & AP$_m$ & AP$_l$ & F & FPS \\
        \hline
        
        \rowcolor{blue} Sparse RCNN~\cite{sun2021sparse} & 100 & 6 & 42.8 & 61.2 & 45.7 & 26.7 & 44.6 & 57.6 & 134 & 26.0 \\
        \rowcolor{blue} Sparse RCNN~\cite{sun2021sparse} & 300 & 6 & 45.0 & 63.4 & 48.2 & 26.9 & 47.2 & 59.5 & 152 & 25.4 \\
        \rowcolor{blue} CF-QRCNN~\cite{zhang2022featurized} & 100 & 2 & 43.0 & 61.3 & 46.8 & 28.3 & 45.7 & 55.5 & 142 & 34.1 \\
        \rowcolor{blue} CF-QRCNN~\cite{zhang2022featurized} & 300 & 2 & 44.6 & 63.1 & 48.9 & 29.5 & 47.4 & 57.5 & 148 & 33.6 \\
        ASAG-S (Ours) & 100 & 1 & 43.9 & 62.2 & 47.9 & 27.8 & 46.5 & 57.8 & 130 & 30.1 \\
        ASAG-S (Ours) & 312 & 1 & 45.0 & 64.1 & 49.1 & 29.5 & 47.4 & 57.8 & 136 & 29.2 \\
        \hline
        \rowcolor{blue} Deform DETR$\ddagger$:~\cite{deformabledetr} & 300 & 6 & 44.5 & 63.6 & 48.7 & 27.1 & 47.6 & 59.6 & 173 & 19.0 \\
        \rowcolor{blue} Deform DETR+$\ddagger$:~\cite{deformabledetr} & 300 & 6 & 46.2 & 64.7 & 49.0 & 28.3 & 49.2 & 61.5 & 173 & 19.0 \\
        ASAG-D (Ours) & 253 & 1 & 45.8 & 64.1 & 49.4 & 27.3 & 49.6 & 61.0 & 182 & 19.7 \\
        \hline
        \rowcolor{blue} AdaMixer*~\cite{adamixer} & 100 & 6 & 45.6 & 64.8 & 49.3 & 28.8 & 48.5 & 60.9 & 103 & 23.8 \\
        \rowcolor{blue} AdaMixer~\cite{adamixer} & 300 & 6 & 47.0 & 66.0 & 51.1 & 30.1 & 50.2 & 61.8 & 125 & 22.2 \\
        \rowcolor{blue} AdaMixer\dag~\cite{adamixer} & 300 & 6 & 48.0 & 67.0 & 52.4 & 30.0 & 51.2 & 63.7 & 201 & 17.6 \\
        ASAG-A (Ours) & 102 & 1 & 45.3 & 63.3 & 48.9 & 27.3 & 48.5 & 59.7 & 131 & 28.9 \\
        ASAG-A (Ours) & 312 & 1 & 46.3 & 65.1 & 50.3 & 29.9 & 49.2 & 59.6 & 139 & 27.9 \\
        ASAG-A (Ours)\dag & 296 & 1 & 47.5 & 66.1 & 51.2 & 30.4 & 50.6 & 62.6 & 206 & 21.3 \\
        \bottomrule
      \end{tabular}
    }
  \vspace{-0.8em}
  \caption{Comparison with six-decoder-layer detectors using 36 epochs and R50. *: Reimplement by us using official codes. \dag: using R101. $\ddagger$: training with 50 epochs.}
  \label{tab:compare_3x}
\end{table}
\subsection{Ablation Studies}
\label{sec:ablation}

\begin{table*}[t]
\centering
\subfloat[
Ablation studies on each component. Replace Anchor denotes replace the selected anchors with newly generated ones in Adaptive Probing.
\label{tab:component}
]{
\centering
\begin{minipage}{0.5\linewidth}{
    \begin{center}
        \tablestyle{1pt}{1.1}
        \begin{tabular}{cccc|ccccccc}
            Dynamic & Query & Auxiliary & Replace & \multirow{2}{*}{AP} & \multirow{2}{*}{AP$_{50}$} & \multirow{2}{*}{AP$_{75}$} & \multirow{2}{*}{AP$_{s}$} & \multirow{2}{*}{AP$_{m}$} & \multirow{2}{*}{AP$_{l}$} & \multirow{2}{*}{\#An} \\
            \#Query & Weighting & Head & Anchor &  &  &  &  &  &  & \\
            \hline
            &  &  & \checkmark & 36.9 & 54.6 & 39.6 & 21.3 & 39.7 & 49.5 & 100 \\
            \checkmark &  &  & \checkmark & 38.9 & 56.7 & 42.1 & 22.7 & 41.7 & 51.5 & 102 \\
            \checkmark & \checkmark &  & \checkmark & 41.6 & 59.2 & 44.9 & 24.5 & 44.3 & 55.6 & 103 \\
            \rowcolor{Gray} \checkmark & \checkmark & \checkmark & \checkmark & 42.6 & 60.5 & 45.8 & 25.9 & 45.8 & 56.9 & 107 \\
            \checkmark & \checkmark & \checkmark &  & 42.4 & 60.2 & 46.0 & 25.2 & 45.5 & 56.9 & 121 \\
        \end{tabular}
    \end{center}}
\end{minipage}
}\hspace{0.5em}
\subfloat[
Ablation studies on \textbf{Confidence Threshold} for Adaptive Probing.
\label{tab:conf_thre}
]{
\begin{minipage}{0.45\linewidth}{
    \begin{center}
        \tablestyle{0.6pt}{1.1}
        \begin{tabular}{c|c|ccccccccc}
            $\eta_h$ & $\eta_l$ & AP & AP$_{50}$ & AP$_{75}$ & AP$_{s}$ & AP$_m$ & AP$_l$ & \#An & N$_{P4}$ & N$_{P3}$ \\
            \hline
            \multirow{4}{*}{0.7} & 0.3 & 41.4 & 58.9 & 44.7 & 22.9 & 45.0 & 57.0 & 99 & 1.3 & 0.6 \\
            & 0.2 & 41.9 & 59.7 & 45.3 & 24.8 & 45.2 & 56.9 & 104 & 2.2 & 1.6 \\
            & 0.1 & \cellcolor{Gray} 42.6 & \cellcolor{Gray} 60.5 & \cellcolor{Gray} 45.8 & \cellcolor{Gray} 25.9 & \cellcolor{Gray} 45.8 & \cellcolor{Gray} 56.9 & \cellcolor{Gray} 107 & \cellcolor{Gray} 4.9 & \cellcolor{Gray} 4.5 \\
            & 0.05 & 42.6 & 60.6 & 46.1 & 25.9 & 45.8 & 56.8 & 124 & 9.9 & 10.9 \\
            \hline
            0.8 & \multirow{4}{*}{0.1}  & 42.5 & 60.4 & 45.8 & 25.9 & 45.7 & 56.9 & 107 & 4.9 & 4.5 \\
            0.7 & & \cellcolor{Gray} 42.6 & \cellcolor{Gray} 60.5 & \cellcolor{Gray} 45.8 & \cellcolor{Gray} 25.9 & \cellcolor{Gray} 45.8 & \cellcolor{Gray} 56.9 & \cellcolor{Gray} 107 & \cellcolor{Gray} 4.9 & \cellcolor{Gray} 4.5 \\
            0.6 &  & 42.5 & 60.4 & 45.8 & 25.9 & 45.7 & 56.9 & 107 &  4.9 & 4.5 \\
            0.4 & & 42.5 & 60.3 & 45.7 & 25.2 & 45.8 & 56.9 & 108 &  4.9 & 4.5 \\
        \end{tabular}
    \end{center}}
\end{minipage}
}\\
\subfloat[
Ablation studies on \textbf{Patch Size} for Adaptive Probing.
\label{tab:pz}
]{
\begin{minipage}{0.33\linewidth}{
    \begin{center}
        \tablestyle{0.6pt}{1.1}
        \begin{tabular}{c|ccccccccc}
            Size & AP & AP$_{50}$ & AP$_{75}$ & AP$_{s}$ & AP$_m$ & AP$_l$ & \#An & N$_{P4}$ & N$_{P3}$ \\
            \hline
            10  & 42.1 & 59.8 & 45.6 & 24.8 & 44.7 & 56.6 & 106 & 4.4 & 3.6 \\
            12  & 42.3 & 60.1 & 45.9 & 25.3 & 45.2 & 56.8 & 108 & 4.9 & 4.3 \\
            \rowcolor{Gray} 15  & 42.6 & 60.5 & 45.8 & 25.9 & 45.8 & 56.9 & 107 & 4.9 & 4.5 \\
            18  & 42.2 & 60.1 & 45.7 & 24.6 & 45.3 & 56.5 & 105 & 4.8 & 4.6 \\
        \end{tabular}
    \end{center}}
\end{minipage}
}\hspace{0.3em}
\subfloat[
Ablation studies on \textbf{NMS Threshold} for Adaptive Probing.
\label{tab:nms}
]{
\begin{minipage}{0.33\linewidth}{
    \begin{center}
        \tablestyle{0.6pt}{1.1}
        \begin{tabular}{c|ccccccccc}
            $\eta_{iou}$ & AP & AP$_{50}$ & AP$_{75}$ & AP$_{s}$ & AP$_m$ & AP$_l$ & \#An & N$_{P4}$ & N$_{P3}$ \\
            \hline
            0.4  & 42.5  & 60.9  & 45.5 & 25.3 & 45.4 & 57.2 & 117 & 7.4 & 7.3 \\
            \rowcolor{Gray} 0.25  & 42.6 & 60.5 & 45.8 & 25.9 & 45.8 & 56.9 & 107 & 4.9 & 4.5 \\
            0.2  & 42.4 & 60.3 & 45.6 & 25.3 & 45.7 & 56.8 & 104 & 4.3 & 3.9 \\
            0.1  & 42.2 & 59.9 & 45.7 & 24.7 & 45.8 & 56.9 & 98 & 3.3 & 2.9 \\
        \end{tabular}
    \end{center}}
\end{minipage}
}\hspace{0.3em}
\subfloat[
Comparison on \textbf{AR} with the 3$\times$ recipe. 
\label{tab:ar}
]{
\centering
\begin{minipage}{0.23\linewidth}{
    \begin{center}
        \tablestyle{0.6pt}{1.1}
          \begin{tabular}{l|cc}
            Method & AR$^{1000}_{100}$ & AR$^{1000}_{300}$ \\
            \hline
            RPN~\cite{ren2015faster} & - & 61.93 \\
            CF-QRCNN~\cite{zhang2022featurized} & 57.31 & 63.42 \\
            \hline
            Dynamic Anchors & 45.22 & 50.21 \\
            Dynamic Proposals & 59.28 & 64.90 \\
          \end{tabular}
    \end{center}}
\end{minipage}
}\\\vspace{-0.5em}
\caption{Experiment results of ASAG-A with the 1$\times$ training recipe and 100 anchors on COCO \texttt{val} except for AR. The settings in our default model are colored in \colorbox{Gray}{gray}. \#An denotes the average number of dynamic anchors. N$_{P4}$ and N$_{P3}$ denote the number of patches used in Adaptive Probing on corresponding feature maps.
}
\label{tab:ablations} \vspace{-0.8em}
\end{table*}

\begin{figure}[t!]
  \centering
  \includegraphics[width=1\linewidth]{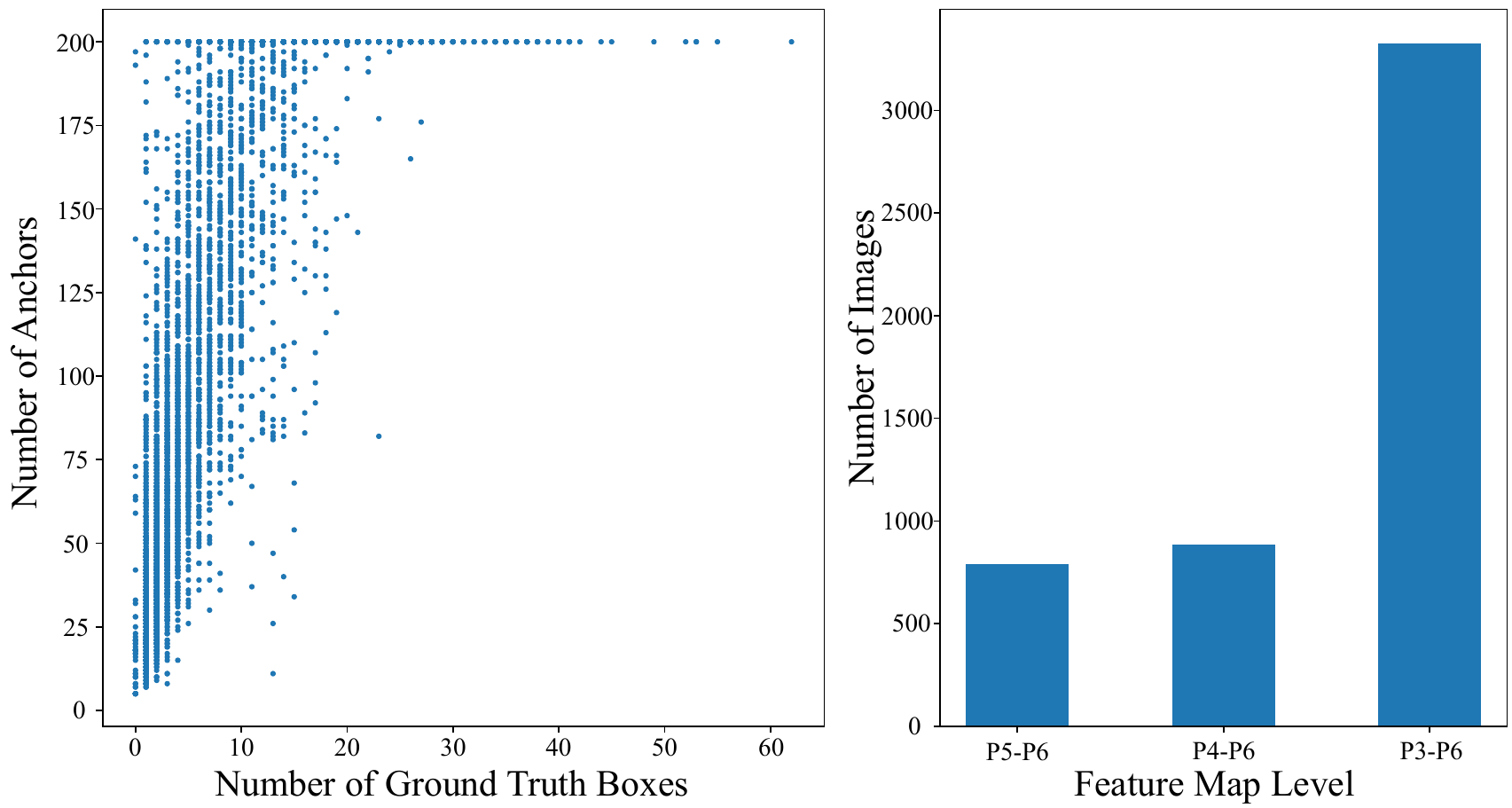}
   \caption{Statistic for Anchor Generator. \textbf{Left}: Correlation between the number of ground truth boxes and the generated anchors. \textbf{Right}: Histogram of used feature map level in Anchor Generator and the number of images.}
   \label{fig:statistic}
\end{figure}

We conduct the following ablation studies with ASAG-A with R50, 100 queries, and 1$\times$ training schedule due to its fast convergence speed.

\vspace{0.5em}\noindent{\textbf{Main components.}} In \Cref{tab:component}, we ablate the main components newly introduced in our model. 
First, since anchors predicted from different feature maps using different predictors can not be treated equally, using dynamic number of anchors rather than fixed topk anchors based on scores is more appropriate, adding 2.0 AP. From the last row, we also find that preserving the selected anchors in Adaptive Probing brings no help but with more anchors, showing that the generated ones on larger feature maps are better than the selected ones and anchors cannot be sorted by scores equally. Second, Query Weighting stabilizes training and brings 2.7 AP gains. Further, providing adequate supervision signals is crucial for sparse detectors and using extra auxiliary parallel one-layer decoders increases 1.0 AP.

\vspace{0.5em}\noindent{\textbf{Patch size for Adaptive Probing.}} Since patches cropped on different feature maps predict anchors independently, patch sizes for each level can be different. For simplicity, we use the same size for each level in Adaptive Probing and share the MLP predictor. In \Cref{tab:pz}, large and small patch sizes are both inappropriate for small objects. For small patch sizes, fewer anchors are selected in Adaptive Probing due to length constraints, resulting in fewer patches. And relatively larger context is needed to detect small and middle objects, similar to the findings in \cite{sa-autoaug}. As for large patch sizes, the predictor overlooks some small objects since we predict sparsely and without using spatial priors.

\vspace{0.5em}\noindent{\textbf{Confidence threshold for Adaptive Probing.}} In the upper part of \Cref{tab:conf_thre}, relatively high $\eta_l$ ignores some possible small objects thus resulting in fewer anchors and patches and poor performance on small objects. Threshold $\eta_l$ lower than 0.1 is not necessary since low confidence anchors tend to be noisy anchors. The lower part shows Adaptive Probing is robust to $\eta_h$. Note that Adaptive Probing will not affect AP$_l$.

\vspace{0.5em}\noindent{\textbf{NMS threshold for Adaptive Probing.}} To reduce redundancy, we use NMS to reduce overlapped patches and the threshold affects the number of patches a lot. In \Cref{tab:nms}, a low threshold reduces some useful patches mistakenly.


\vspace{0.5em}\noindent{\textbf{Comparison on AR.}} To validate the quality of our dynamic anchors and dynamic proposals, in~\Cref{tab:ar}, we compare AR$^{1000}$ with well-known RPN~\cite{ren2015faster} and QGN used in~\cite{zhang2022featurized}. AR$^{1000}_{100}$ means AR$^{1000}$ under the 100 queries setting. Although our model lacks dense priors, like anchor boxes or anchor points, dynamic proposals still outperform RPN and QGN in terms of AR$^{1000}$. And our Anchor Generator is lightweight using only 0.06 GFLOPs.

\vspace{0.5em}\noindent{\textbf{Distribution of the number of dynamic anchors.}} As shown in the left part of \Cref{fig:statistic}, there is a clear positive correlation between the number of ground truth boxes and the generated anchors, showing that the Anchor Generator is adaptive to different images by generating more queries for difficult images and vice versa.

\vspace{0.5em}\noindent{\textbf{Distribution of the number of used feature map levels.}} Our Adaptive Probing method enjoys the early-stop mechanism. As shown in the right part of \Cref{fig:statistic}, roughly 40\% of the images in the validation set do not use all the feature maps, saving some computation.

\begin{table}[t]
  \centering
  \resizebox{\linewidth}{!}{
      \begin{tabular}{c|c|ccc|c|ccc}
        \hline
        $\eta_h$ & $\eta_l$ & AP$(\uparrow)$ & mMR$(\downarrow)$ & R$(\uparrow)$ & $\eta_{iou}$ & AP$(\uparrow)$ & mMR$(\downarrow)$ & R$(\uparrow)$ \\
        \hline
        \multirow{4}{*}{0.7} & 0.3 & 90.8 & 44.0 & 96.4 & 0.4 & 90.6 & 43.7 & 96.0 \\
        & 0.2 & 91.2 & 43.7 & 96.7 & \cellcolor{Gray} 0.25 & \cellcolor{Gray} 91.3 & \cellcolor{Gray} 43.5 & \cellcolor{Gray} 96.9 \\
        & 0.1 & \cellcolor{Gray} 91.3 & \cellcolor{Gray} 43.5 & \cellcolor{Gray} 96.9 & 0.2 & 91.2 & 43.8 & 96.8  \\
        & 0.05 & 91.4 & 43.5 & 96.9 & 0.1 & 90.2 & 44.4 & 95.6 \\
        \hline
        0.8 & \multirow{4}{*}{0.1}  & 91.3 & 43.5 & 96.9 & Sparse* & \multirow{2}{*}{89.2} & \multirow{2}{*}{48.3} & \multirow{2}{*}{95.9} \\
        0.7 & & \cellcolor{Gray} 91.3 & \cellcolor{Gray} 43.5 & \cellcolor{Gray} 96.9 & RCNN \\
        \cline{6-9}
        0.6 &  & 91.3 & 43.5 & 96.9 & Deformable* & \multirow{2}{*}{86.7} & \multirow{2}{*}{54.0} & \multirow{2}{*}{92.5} \\
        0.5 & & 91.2 & 43.6 & 96.7 & DETR \\ \hline
      \end{tabular}
    }
  \caption{CrowdHuman results on different \textbf{Confidence Threshold}s and \textbf{NMS Threshold}s for Adaptive Probing using ASAG-S. Rows in \colorbox{Gray}{gray} denote the default settings on COCO. *: Results are taken from Sparse RCNN\cite{sun2021sparse}.}
  \label{tab:crowdhuman}
\end{table}

\vspace{0.5em}\noindent{\textbf{Robustness of hyper-parameters in Adaptive Probing.}} In Adaptive Probing, we only select anchors whose confidences fall in [$\eta_l$, $\eta_h$] to crop patches in the larger feature maps and the patches are filtered by NMS with IoU threshold $\eta_{iou}$ for reducing redundancy. To show the robustness of these hyper-parameters, we follow Sparse RCNN~\cite{sun2021sparse} to conduct experiments on CrowdHuman~\cite{shao2018crowdhuman} dataset, which is significantly different from COCO. Following Sparse RCNN, we run ASAG-S with 50 epochs and the average number of anchors within 500. As shown in \Cref{tab:crowdhuman}, the default hyper-parameters of Adaptive Probing on COCO still work on CrowdHuman. And ASAG-S outperforms Sparse RCNN and Deformable DETR by a large margin.

\subsection{Visualization}
\label{sec:visualization}
\begin{figure}[t!]
  \centering
  \includegraphics[width=0.99\linewidth]{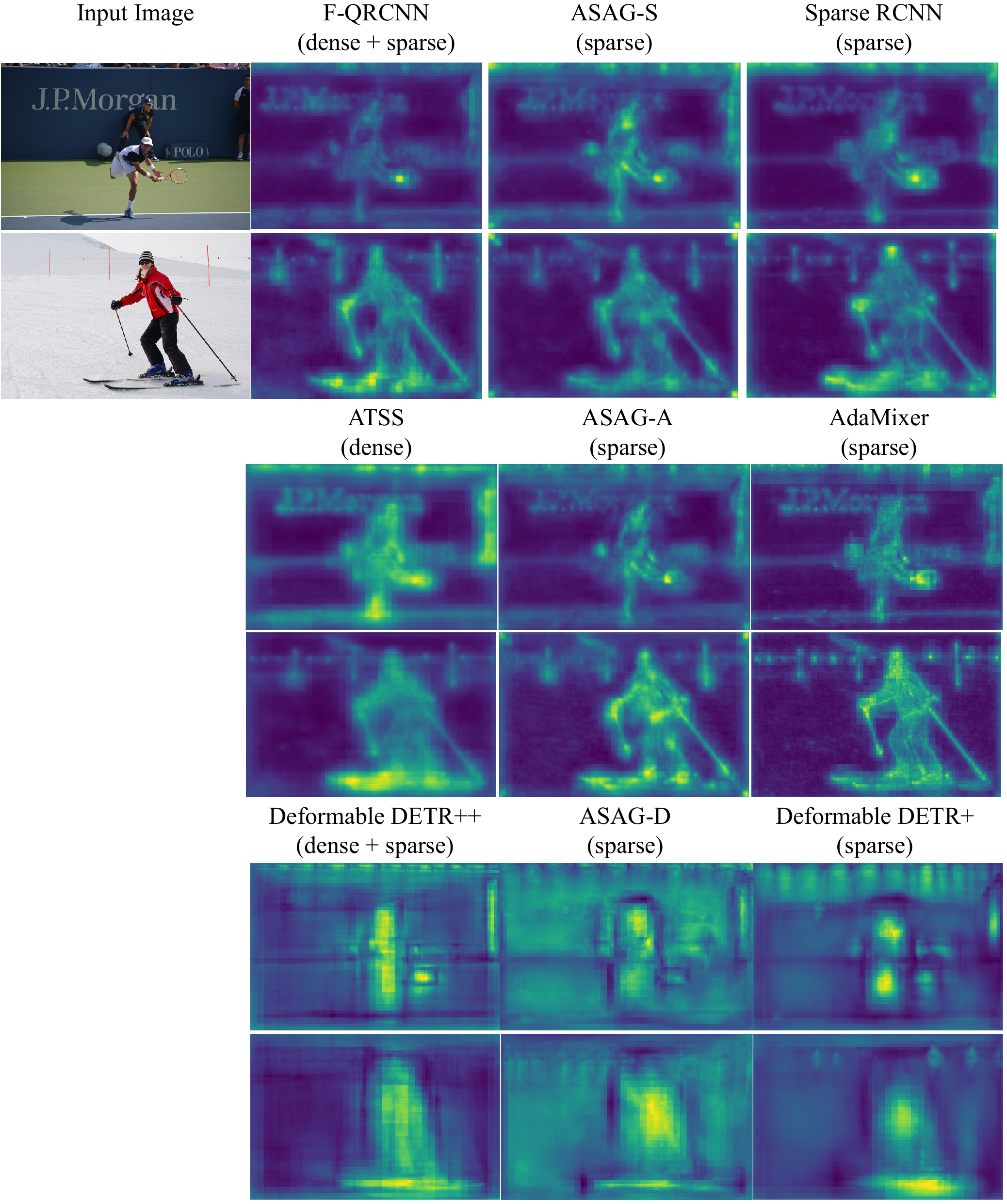}
  \vspace{-0.2em}
   \caption{Visualization of feature maps of different methods. Our models with sparse initialization are more consistent with sparse decoders.}
   \label{fig:experiment_feature}
\end{figure}

\noindent{\textbf{Comparison on feature maps.}} In this work, we provide a sparse way to initialize object queries by using patches as the prediction units thus alleviating the feature map discrepancy caused by predicting on grids. As shown in \Cref{fig:experiment_feature}, feature maps from our method are more similar to six-decoder-layer sparse detectors, which activate objects in an adaptive way rather than uniformly.

\begin{figure}[t!]
  \centering
  \includegraphics[width=0.99\linewidth]{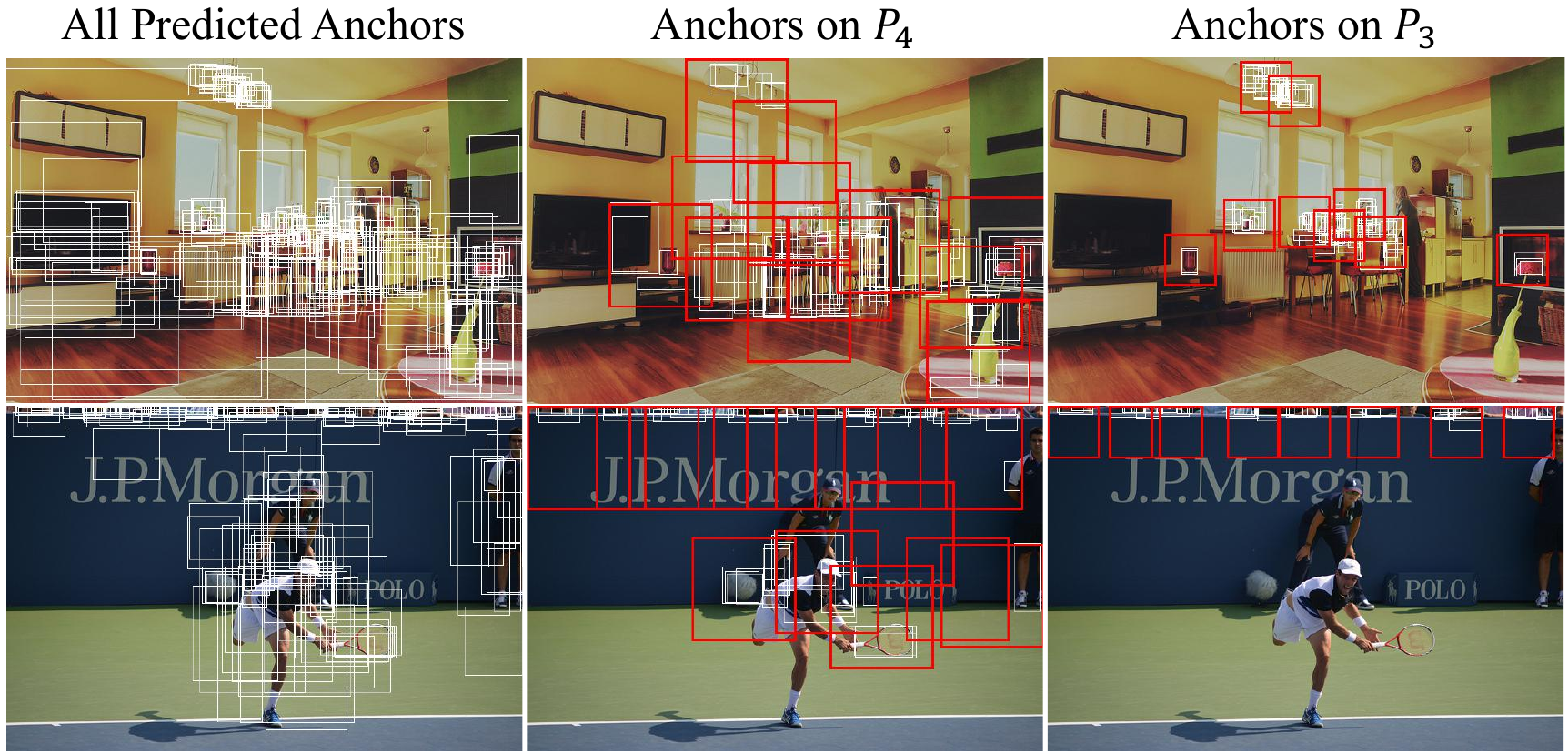}
  \vspace{-0.2em}
   \caption{Visualization of dynamic anchors and Adaptive Probing. The white and red boxes represent anchors and patches, respectively.}
   \label{fig:anchors}
\end{figure}

\vspace{0.5em}\noindent{\textbf{Visualization for Adaptive Probing.}} We visualize some results to understand how Anchor Generator works. More pictures will be displayed in the supplementary materials. In \Cref{fig:anchors}, we draw dynamic anchors in white and patches in red. As shown in the first column, the anchors for different images are different and have covered most foreground objects, showing them truly adaptive and precise. Anchor Generator tends to generate more anchors for small objects through Adaptive Probing, which greatly enhances the ability to detect small objects. Although the red boxes, \ie patches, sparsely locate on the images, they precisely cover small objects, such as the things on the table and the spectators in the stands, greatly increasing the recall rate.

\subsection{Discussion}
\label{sec:discussion}

\begin{table}[t]
  \centering
  \resizebox{0.98\linewidth}{!}{
      \begin{tabular}{l|c|cccccc}
        \hline
        Detector & Query Weighting & AP & AP$_{50}$ & AP$_{75}$ & AP$_{s}$ & AP$_m$ & AP$_l$ \\
        \hline
        \multirow{2}{*}{AdaMixer~\cite{adamixer}} & \checkmark & 39.2 & 47.7 & 42.1 & 21.5 & 42.2 & 55.3  \\
         & & 42.7 & 61.5 & 45.9 & 24.7 & 45.4 & 59.2 \\
        \hline
      \end{tabular}
    }
    \vspace{-0.4em}
  \caption{Query Weighting for six-decoder-layer detectors.}
  \label{tab:discussion}
\end{table}

\noindent\textbf{Is Query Weighting beneficial to six-decoder-layer detectors?} Most six-decoder-layer detectors and dense ones detect objects from image-agnostic and fixed queries (or anchors). The quality of initial queries can not be improved during the training process. Thus each decoder layer should be trained under difference IoU level~\cite{cascade_rcnn} and Query Weighting can not benefit them, as shown in \Cref{tab:discussion}.

\vspace{0.5em}\noindent\textbf{Is it equal to increase the same number of anchors for one- and six-decoder-layer detectors?} Since six-decoder-layer detectors work well with image-agnostic and fixed queries, query initialization brings minor help to them. But more queries for them narrow the search space for each query. However, query initialization is vital to one-decoder-layer detectors and directly affects their performance~\cite{efficient_detr}. As shown in \Cref{tab:ar}, increasing queries from 100 to 300 only brings 5.6 AR since the top100 proposals already lie in the top300 ones. Thus the benefit of increasing queries for one-decoder-layer detectors is less than for six-decoder-layer ones, shown in \Cref{tab:compare_1x} and \Cref{tab:compare_3x}. How to scale up one-decoder-layer detectors needs future work.

\vspace{0.5em}\noindent\textbf{More analysis on AP$_l$.} A core advantage of DETR-like detectors is that they reason globally so that the AP$_l$ is significantly higher than traditional detectors. However, although our ASAGs are comparable on AP with corresponding six-decoder-layer counterparts, the AP$_l$ still lags behind them, as shown in \Cref{tab:compare_3x}. And it seems that the AP$_l$ of ASAGs cannot increase when using more queries just as the counterparts. Here we provide some analysis.

The AP$_l$ is affected by two factors: the number of decoder layers and query initialization. Regarding the number of decoder layers, we run ASAG-A with two decoder layers in \Cref{tab:more_layer} and get 43.2AP and 59.0AP$_l$ (vs one layer 42.6AP and 56.9AP$_l$), in which AP$_l$ is already comparable with AdaMixer. As for query initialization, ASAG views the whole image in $P_6$ as a patch and predicts anchors from it, enjoying global receptive fields during initialization, and thus shows superiority to dense-initialized one-decoder-layer detectors on AP$_l$ in \Cref{tab:compare_one_layer_3x}. 

\begin{table}[t]
  \centering
  \resizebox{0.98\linewidth}{!}{
      \begin{tabular}{c|c|cccccc}
        \hline
        Detector & \#L & AP & AP$_{50}$ & AP$_{75}$ & AP$_{s}$ & AP$_m$ & AP$_l$ \\
        \hline
        ASAG-A & 1 & 42.6 & 60.5 & 45.8 & 25.9 & 45.8 & 56.9  \\
        ASAG-A & 2 & 43.2 & 61.6 & 46.9 & 25.6 & 46.4 & 59.0 \\
        AdaMixer & 6 & 42.7 & 61.5 & 45.9 & 24.7 & 45.4 & 59.2 \\
        \hline
      \end{tabular}
    }
    \vspace{-0.2em}
  \caption{Effect of different number of decoder layers.}
  \label{tab:more_layer}
\end{table}

Besides, the phenomenon that AP$_l$ does not improve with more queries also occurred in another one-decoder-layer detector (see F-QRCNN in \Cref{tab:compare_one_layer_3x}), showing it is not specific to ASAG. Since one-decoder-layer detectors use image-specific queries and large objects are relatively easy to detect, the initial queries are accurate enough, and increasing queries brings a little recall for large objects. In contrast, queries of six-decoder-layer detectors are randomly initialized. Thus, the gains on AP$_l$ for one-decoder-layer detectors are smaller than six-decoder-layer counterparts when using more queries.

\section{Conclusion}
In this work, we find that dense initialization is not optimal for one-decoder-layer sparse detectors since predicting on grids leads to feature conflict, which hampers their performance. To tackle this problem, we propose ASAG and predict dynamic anchors based on patches in a sparse way, thus alleviating feature conflict. We further design Adaptive Probing to generate patches on different levels, which is adaptive to the number of used feature maps, the number of patches, and the locations of patches. Finally, simple but effective Query Weighting stabilizes the training. With the novel design, our dynamic anchors and proposals are better than dense ones without using predefined spatial priors. Experiments show that we greatly increase the performance of one-decoder-layer detectors and narrow the performance gap while retaining the fast speed.

\vspace{0.5em}\noindent{\textbf{Acknowledgments.}} 
This work was supported partially by the NSFC (U21A20471, U1911401, 62072482), Guangdong NSF Project (No. 2023B1515040025, 2020B1515120085).

{\small
\bibliographystyle{ieee_fullname}
\bibliography{egbib}

\begin{thebibliography}{10}\itemsep=-1pt

\bibitem{cascade_rcnn}
Zhaowei Cai and Nuno Vasconcelos.
\newblock Cascade r-cnn: Delving into high quality object detection.
\newblock In {\em CVPR}, 2018.

\bibitem{detr}
Nicolas Carion, Francisco Massa, Gabriel Synnaeve, Nicolas Usunier, Alexander
  Kirillov, and Sergey Zagoruyko.
\newblock End-to-end object detection with transformers.
\newblock In {\em ECCV}, 2020.

\bibitem{detrdistill}
Jiahao Chang, Shuo Wang, Guangkai Xu, Zehui Chen, Chenhongyi Yang, and Feng
  Zhao.
\newblock Detrdistill: A universal knowledge distillation framework for
  detr-families.
\newblock {\em arXiv preprint arXiv:2211.10156}, 2022.

\bibitem{group_detr}
Qiang Chen, Xiaokang Chen, Gang Zeng, and Jingdong Wang.
\newblock Group detr: Fast training convergence with decoupled one-to-many
  label assignment.
\newblock {\em arXiv preprint arXiv:2207.13085}, 2022.

\bibitem{d3etr}
Xiaokang Chen, Jiahui Chen, Yan Liu, and Gang Zeng.
\newblock D3etr: Decoder distillation for detection transformer.
\newblock {\em arXiv preprint arXiv:2211.09768}, 2022.

\bibitem{sa-autoaug}
Yukang Chen, Yanwei Li, Tao Kong, Lu Qi, Ruihang Chu, Lei Li, and Jiaya Jia.
\newblock Scale-aware automatic augmentation for object detection.
\newblock In {\em CVPR}, 2021.

\bibitem{REGO}
Zhe Chen, Jing Zhang, and Dacheng Tao.
\newblock Recurrent glimpse-based decoder for detection with transformer.
\newblock In {\em CVPR}, 2022.

\bibitem{tood}
Chengjian Feng, Yujie Zhong, Yu Gao, Matthew~R Scott, and Weilin Huang.
\newblock Tood: Task-aligned one-stage object detection.
\newblock In {\em ICCV}, 2021.

\bibitem{smca_detr}
Peng Gao, Minghang Zheng, Xiaogang Wang, Jifeng Dai, and Hongsheng Li.
\newblock Fast convergence of detr with spatially modulated co-attention.
\newblock In {\em ICCV}, 2021.

\bibitem{adamixer}
Ziteng Gao, Limin Wang, Bing Han, and Sheng Guo.
\newblock Adamixer: A fast-converging query-based object detector.
\newblock In {\em CVPR}, 2022.

\bibitem{he2017mask}
Kaiming He, Georgia Gkioxari, Piotr Doll{\'a}r, and Ross Girshick.
\newblock Mask r-cnn.
\newblock In {\em ICCV}, 2017.

\bibitem{hong2022dynamic}
Qinghang Hong, Fengming Liu, Dong Li, Ji Liu, Lu Tian, and Yi Shan.
\newblock Dynamic sparse r-cnn.
\newblock In {\em CVPR}, 2022.

\bibitem{teach_detr}
Linjiang Huang, Kaixin Lu, Guanglu Song, Liang Wang, Si Liu, Yu Liu, and
  Hongsheng Li.
\newblock Teach-detr: Better training detr with teachers.
\newblock {\em arXiv preprint arXiv:2211.11953}, 2022.

\bibitem{jia2022detrs}
Ding Jia, Yuhui Yuan, Haodi He, Xiaopei Wu, Haojun Yu, Weihong Lin, Lei Sun,
  Chao Zhang, and Han Hu.
\newblock Detrs with hybrid matching.
\newblock {\em arXiv preprint arXiv:2207.13080}, 2022.

\bibitem{h_detr}
Ding Jia, Yuhui Yuan, Haodi He, Xiaopei Wu, Haojun Yu, Weihong Lin, Lei Sun,
  Chao Zhang, and Han Hu.
\newblock Detrs with hybrid matching.
\newblock {\em arXiv preprint arXiv:2207.13080}, 2022.

\bibitem{pointrend}
Alexander Kirillov, Yuxin Wu, Kaiming He, and Ross Girshick.
\newblock Pointrend: Image segmentation as rendering.
\newblock In {\em CVPR}, 2020.

\bibitem{dn_detr}
Feng Li, Hao Zhang, Shilong Liu, Jian Guo, Lionel~M Ni, and Lei Zhang.
\newblock Dn-detr: Accelerate detr training by introducing query denoising.
\newblock In {\em CVPR}, 2022.

\bibitem{dw}
Shuai Li, Chenhang He, Ruihuang Li, and Lei Zhang.
\newblock A dual weighting label assignment scheme for object detection.
\newblock In {\em CVPR}, 2022.

\bibitem{gfl}
Xiang Li, Wenhai Wang, Lijun Wu, Shuo Chen, Xiaolin Hu, Jun Li, Jinhui Tang,
  and Jian Yang.
\newblock Generalized focal loss: Learning qualified and distributed bounding
  boxes for dense object detection.
\newblock In {\em NeurIPS}, 2020.

\bibitem{dpp}
Yunsheng Li, Yinpeng Chen, Xiyang Dai, Dongdong Chen, Mengchen Liu, Pei Yu,
  Ying Jin, Lu Yuan, Zicheng Liu, and Nuno Vasconcelos.
\newblock Should all proposals be treated equally in object detection?
\newblock In {\em ECCV}, 2022.

\bibitem{lin2017feature}
Tsung-Yi Lin, Piotr Doll{\'a}r, Ross Girshick, Kaiming He, Bharath Hariharan,
  and Serge Belongie.
\newblock Feature pyramid networks for object detection.
\newblock In {\em CVPR}, 2017.

\bibitem{lin2017focal}
Tsung-Yi Lin, Priya Goyal, Ross Girshick, Kaiming He, and Piotr Doll{\'a}r.
\newblock Focal loss for dense object detection.
\newblock In {\em ICCV}, 2017.

\bibitem{coco}
Tsung-Yi Lin, Michael Maire, Serge Belongie, James Hays, Pietro Perona, Deva
  Ramanan, Piotr Doll{\'a}r, and C~Lawrence Zitnick.
\newblock Microsoft coco: Common objects in context.
\newblock In {\em ECCV}, 2014.

\bibitem{dab_detr}
Shilong Liu, Feng Li, Hao Zhang, Xiao Yang, Xianbiao Qi, Hang Su, Jun Zhu, and
  Lei Zhang.
\newblock Dab-detr: Dynamic anchor boxes are better queries for detr.
\newblock In {\em ICLR}, 2022.

\bibitem{SAP-detr}
Yang Liu, Yao Zhang, Yixin Wang, Yang Zhang, Jiang Tian, Zhongchao Shi,
  Jianping Fan, and Zhiqiang He.
\newblock Sap-detr: Bridging the gap between salient points and queries-based
  transformer detector for fast model convergency.
\newblock {\em arXiv preprint arXiv:2211.02006}, 2022.

\bibitem{adamw}
Ilya Loshchilov and Frank Hutter.
\newblock Decoupled weight decay regularization.
\newblock In {\em ICLR}, 2019.

\bibitem{conditional_detr}
Depu Meng, Xiaokang Chen, Zejia Fan, Gang Zeng, Houqiang Li, Yuhui Yuan, Lei
  Sun, and Jingdong Wang.
\newblock Conditional detr for fast training convergence.
\newblock In {\em ICCV}, 2021.

\bibitem{nms_strikes_back}
Jeffrey Ouyang-Zhang, Jang~Hyun Cho, Xingyi Zhou, and Philipp
  Kr{\"a}henb{\"u}hl.
\newblock Nms strikes back.
\newblock {\em arXiv preprint arXiv:2212.06137}, 2022.

\bibitem{ren2015faster}
Shaoqing Ren, Kaiming He, Ross Girshick, and Jian Sun.
\newblock Faster r-cnn: Towards real-time object detection with region proposal
  networks.
\newblock In {\em NeurIPS}, 2015.

\bibitem{giou}
Hamid Rezatofighi, Nathan Tsoi, JunYoung Gwak, Amir Sadeghian, Ian Reid, and
  Silvio Savarese.
\newblock Generalized intersection over union: A metric and a loss for bounding
  box regression.
\newblock In {\em CVPR}, 2019.

\bibitem{sparse_detr}
Byungseok Roh, JaeWoong Shin, Wuhyun Shin, and Saehoon Kim.
\newblock Sparse detr: Efficient end-to-end object detection with learnable
  sparsity.
\newblock In {\em ICLR}, 2022.

\bibitem{shao2018crowdhuman}
Shuai Shao, Zijian Zhao, Boxun Li, Tete Xiao, Gang Yu, Xiangyu Zhang, and Jian
  Sun.
\newblock Crowdhuman: A benchmark for detecting human in a crowd.
\newblock {\em arXiv preprint arXiv:1805.00123}, 2018.

\bibitem{sun2021sparse}
Peize Sun, Rufeng Zhang, Yi Jiang, Tao Kong, Chenfeng Xu, Wei Zhan, Masayoshi
  Tomizuka, Lei Li, Zehuan Yuan, Changhu Wang, et~al.
\newblock Sparse r-cnn: End-to-end object detection with learnable proposals.
\newblock In {\em CVPR}, 2021.

\bibitem{tian2019fcos}
Zhi Tian, Chunhua Shen, Hao Chen, and Tong He.
\newblock Fcos: Fully convolutional one-stage object detection.
\newblock In {\em ICCV}, 2019.

\bibitem{pnp_detr}
Tao Wang, Li Yuan, Yunpeng Chen, Jiashi Feng, and Shuicheng Yan.
\newblock Pnp-detr: towards efficient visual analysis with transformers.
\newblock In {\em ICCV}, 2021.

\bibitem{anchor_detr}
Yingming Wang, Xiangyu Zhang, Tong Yang, and Jian Sun.
\newblock Anchor detr: Query design for transformer-based detector.
\newblock In {\em AAAI}, 2022.

\bibitem{yang2022querydet}
Chenhongyi Yang, Zehao Huang, and Naiyan Wang.
\newblock Querydet: Cascaded sparse query for accelerating high-resolution
  small object detection.
\newblock In {\em CVPR}, 2022.

\bibitem{efficient_detr}
Zhuyu Yao, Jiangbo Ai, Boxun Li, and Chi Zhang.
\newblock Efficient detr: improving end-to-end object detector with dense
  prior.
\newblock {\em arXiv preprint arXiv:2104.01318}, 2021.

\bibitem{sam_detr}
Gongjie Zhang, Zhipeng Luo, Yingchen Yu, Kaiwen Cui, and Shijian Lu.
\newblock Accelerating detr convergence via semantic-aligned matching.
\newblock In {\em CVPR}, 2022.

\bibitem{sam_detr++}
Gongjie Zhang, Zhipeng Luo, Yingchen Yu, Jiaxing Huang, Kaiwen Cui, Shijian Lu,
  and Eric~P Xing.
\newblock Semantic-aligned matching for enhanced detr convergence and
  multi-scale feature fusion.
\newblock {\em arXiv preprint arXiv:2207.14172}, 2022.

\bibitem{IMFA}
Gongjie Zhang, Zhipeng Luo, Yingchen Yu, Zichen Tian, Jingyi Zhang, and Shijian
  Lu.
\newblock Towards efficient use of multi-scale features in transformer-based
  object detectors.
\newblock {\em arXiv preprint arXiv:2208.11356}, 2022.

\bibitem{zhang2022dino}
Hao Zhang, Feng Li, Shilong Liu, Lei Zhang, Hang Su, Jun Zhu, Lionel~M Ni, and
  Heung-Yeung Shum.
\newblock Dino: Detr with improved denoising anchor boxes for end-to-end object
  detection.
\newblock {\em arXiv preprint arXiv:2203.03605}, 2022.

\bibitem{atss}
Shifeng Zhang, Cheng Chi, Yongqiang Yao, Zhen Lei, and Stan~Z Li.
\newblock Bridging the gap between anchor-based and anchor-free detection via
  adaptive training sample selection.
\newblock In {\em CVPR}, 2020.

\bibitem{zhang2022featurized}
Wenqiang Zhang, Tianheng Cheng, Xinggang Wang, Qian Zhang, and Wenyu Liu.
\newblock Featurized query r-cnn.
\newblock {\em arXiv preprint arXiv:2206.06258}, 2022.

\bibitem{act}
Minghang Zheng, Peng Gao, Renrui Zhang, Kunchang Li, Xiaogang Wang, Hongsheng
  Li, and Hao Dong.
\newblock End-to-end object detection with adaptive clustering transformer.
\newblock In {\em BMVC}, 2021.

\bibitem{deformabledetr}
Xizhou Zhu, Weijie Su, Lewei Lu, Bin Li, Xiaogang Wang, and Jifeng Dai.
\newblock Deformable detr: Deformable transformers for end-to-end object
  detection.
\newblock In {\em ICLR}, 2021.

\bibitem{co_detr}
Zhuofan Zong, Guanglu Song, and Yu Liu.
\newblock Detrs with collaborative hybrid assignments training.
\newblock {\em arXiv preprint arXiv:2211.12860}, 2022.

\end{thebibliography}
}

\newpage
\mbox{}
\newpage

\begin{appendices}
\setcounter{table}{0} 
\setcounter{figure}{0} 
\setcounter{equation}{0} 
\renewcommand{\thetable}{\thesection-\arabic{table}} 
\renewcommand{\theequation}{\thesection-\arabic{equation}} 
\renewcommand{\thefigure}{\thesection-\arabic{figure}} 
\setcounter{table}{0} 
\setcounter{figure}{0} 
\setcounter{equation}{0} 

\section{More Details about Loss Function}
In this work, we use patches as the basic prediction units in Anchor Generator. We compute bipartite matching and losses for each patch independently and the targets for each patch are objects whose centers lie in the patch.

\begin{figure}[h]
  \centering
  \includegraphics[width=0.8\linewidth]{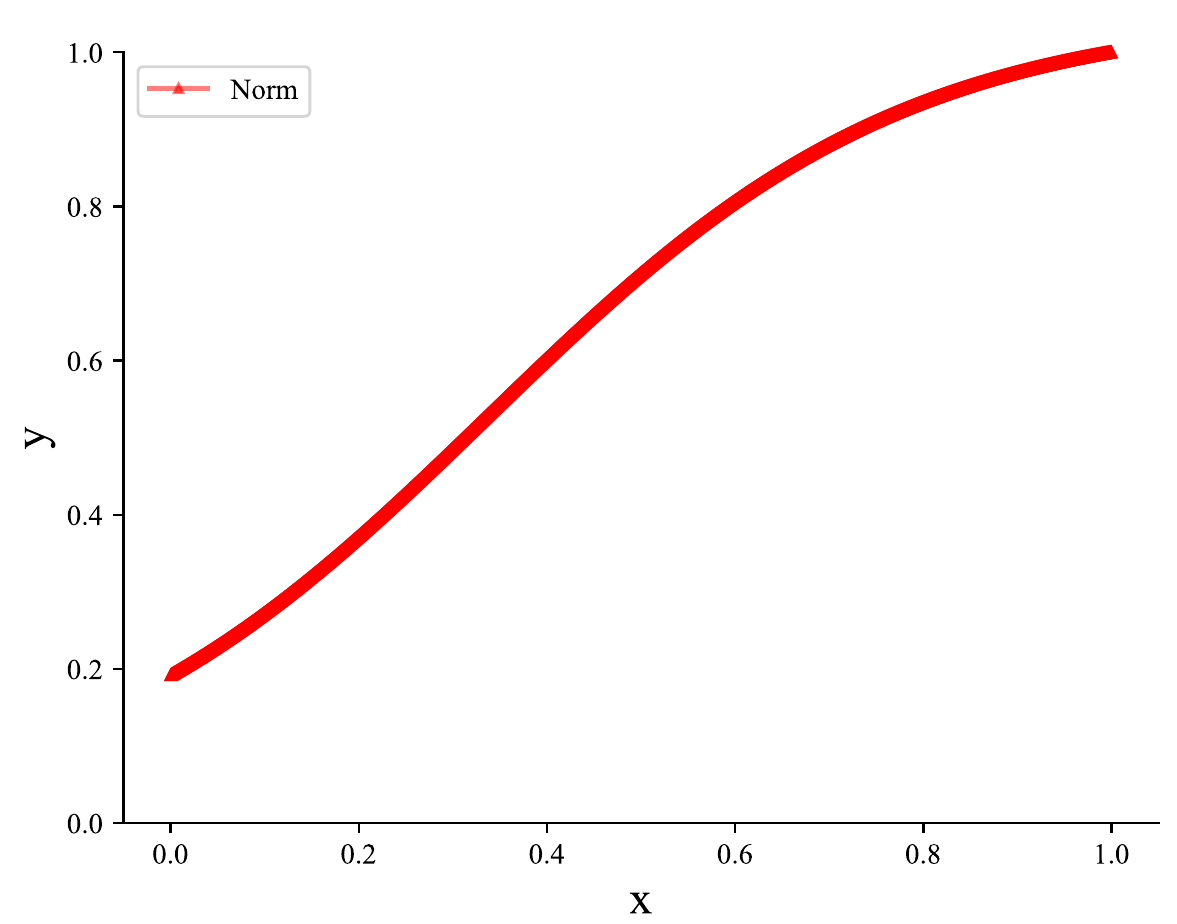}
  \caption{Visualization of the normalization function in Query Weighting.}
  \label{fig:norm}
\end{figure}

Further, we propose Query Weighting to stabilize the training process, which gives high-quality anchors with larger weights and vice versa. The \texttt{Norm} function is shown in \Cref{fig:norm}. The variable $x$ in the picture is the product of $x_1$ and $x_2$ in Equ. (1) of the main text. The monotonically increasing normalization function raises small values and keeps them smaller than 1.

Following other DETR-like models, we use L1 loss and GIoU loss~\cite{giou} with Query Weighting for box regression:
\begin{equation}
\begin{aligned}
    \mathcal{L}_{box}(\hat{b}, b) = &\lambda_1\times w_{pos}\times \mathcal{L}_{L1}(\hat{b}, b) \\
    &+ \lambda_2\times w_{pos}\times \mathcal{L}_{GIoU}(\hat{b}, b),
    \label{eq:box_loss}
\end{aligned}
\end{equation}
where $\hat{b}$ and $b$ are the ground truth and the predicted box, respectively. The $\lambda_1$ and $\lambda_2$ are set to 5 and 2. $w_{pos}$ is defined in Equ. (2) in the main text. The classification loss for negative samples is sigmoid focal loss~\cite{lin2017focal} and the classification loss for positive samples is defined as follows:
\begin{equation}
    \mathcal{L}_{cls}(s) = -\lambda_3\times (w_{pos}\times \log{s} + w_{neg}\times \log{(1-s)}),
    \label{eq:class_loss}
\end{equation}
where $s$ is the classification score with respect to the corresponding class and $\lambda_3$ is set to 2. $w_{neg}$ is defined in Equ. (3) in the main text. In particular, $w_{pos}$ in classification loss for Anchor Generator is set to IoU as dynamic anchors are class-agnostic and the location scores should be highly correlated to IoUs for selection. The overall losses are the sum of all components:
\begin{equation}
    \mathcal{L}_{all} = \lambda_{an}\mathcal{L}_{anchor} + \mathcal{L}_{proposal} + \mathcal{L}_{final} + \sum_{i=0}^2\mathcal{L}_{auxiliary}^i,
    \label{eq:all_loss}
\end{equation}

\begin{table}[t]
  \centering
  \resizebox{0.98\linewidth}{!}{
      \begin{tabular}{c|cccccc}
        \hline
        Denoising Training & AP & AP$_{50}$ & AP$_{75}$ & AP$_{s}$ & AP$_m$ & AP$_l$ \\
        \hline
         & 42.6 & 60.5 & 45.8 & 25.9 & 45.8 & 56.9  \\
        \checkmark & 43.1 & 60.2 & 46.7 & 25.1 & 45.8 & 58.4 \\
        \hline
      \end{tabular}
    }
    \vspace{-0.2em}
  \caption{Equipping ASAG-A with Denoising Training.}
  \label{tab:dn}
\end{table}

Different from losses, the matching cost in bipartite matching does not use Query Weighting.

\begin{table*}[t]
\centering
\scalebox{0.91}{
    
    \small
    \renewcommand\arraystretch{1.0}
    \setlength{\tabcolsep}{4pt}
    \begin{tabular}{l|c|c|c|c|llllll}
        \hline
        Detector & Backbone & {\#Layers} & {\#Epochs} & { GFLOPs} & AP  & AP$_{50}$ & AP$_{75}$ & AP$_{s}$ & AP$_m$ & AP$_l$  \\
        \hline
        DETR~\cite{detr}
        & ResNet-50-DC5 & 6 & 500 & 187 & 43.3 & 63.1 & 45.9 & 22.5 & 47.3 & 61.1  \\
        SMCA~\cite{smca_detr}
        & ResNet-50 & 6 & 50 & 152 & 43.7 & 63.6 & 47.2 & 24.2 & 47.0 & 60.4 \\
        Deformable DETR~\cite{deformabledetr}
        & ResNet-50 & 6 & 50 & 173 & 43.8 & 62.6 & 47.7 & 26.4 & 47.1 & 58.0  \\
        Sparse RCNN~\cite{sun2021sparse}
        & ResNet-50 & 6 & 36 & 152  & 45.0 & 63.4 & 48.2 & 26.9 & 47.2 & 59.5  \\
        Dynamic Sparse RCNN~\cite{hong2022dynamic}
        & ResNet-50 & 6 & 36 & -  & 47.2 & 66.5 & 51.2 & 30.1 & 50.4 & 61.7  \\
        Conditional DETR~\cite{conditional_detr}
        & ResNet-50-DC5 & 6 & 108 & 195 & 45.1 & 65.4 & 48.5 & 25.3 & 49.0 & 62.2\\
        Anchor DETR~\cite{anchor_detr}
        & ResNet-50-DC5 & 6 & 50 & 151 & 44.2 & 64.7 & 47.5 & 24.7 & 48.2 & 60.6 \\
        DAB-DETR~\cite{dab_detr}
        & ResNet-50-DC5 & 6 & 50 & 202 & 44.5 & 65.1 & 47.7 & 25.3 & 48.2 & 62.3  \\
        DN-DETR~\cite{dn_detr}
        & ResNet-50-DC5 & 6 & 50 & 202 & 46.3 & 66.4 & 49.7 & 26.7 & 50.0 & 64.3  \\
        SAM-DETR-R50 w/ SMCA~\cite{sam_detr}
        & ResNet-50-DC5 & 6 & 50 & 210 & 45.0 & 65.4 & 47.9 & 26.2 & 49.0 & 63.3  \\
        DINO-4scale~\cite{zhang2022dino}
        & ResNet-50 & 6 & 24 & 279 & 49.9 & 67.4 & 54.5 & 31.8 & 53.3 & 64.3  \\
        AdaMixer~\cite{adamixer}
        & ResNet-50 & 6 & 36 & 125 & 47.0 & 66.0 & 51.1 & 30.1 & 50.2 & 61.8  \\
        DAB-DETR-R50 + IMFA~\cite{IMFA}
        & ResNet-50 & 6 & 50 & 108 & 45.5 & 65.0 & 49.3 & 27.3 & 48.3 & 61.6  \\
        REGO-Deformable DETR~\cite{REGO}
        & ResNet-50 & 12 & 50 & 190 & 47.6 & 66.8 & 51.6 & 29.6 & 50.6 & 62.3  \\
        SAP-DETR-DC5~\cite{SAP-detr}
        & ResNet-50-DC5 & 6 & 50 & 197 & 46.0 & 65.5 & 48.9 & 26.4 & 50.2 & 62.6  \\
        Efficient DETR~\cite{efficient_detr}
        & ResNet-50 & 1 & 36 & 210 & 45.1 & 63.1 & 49.1 & 28.3 & 48.4 & 59.0 \\
        Cascade Featurized QRCNN~\cite{zhang2022featurized}
        & ResNet-50 & 2 & 36 & 148 & 44.6 & 63.1 & 48.9 & 29.5 & 47.4 & 57.5  \\
        \hline
        ASAG-S (Ours)
        & ResNet-50 & 1 & 36 & 136 & 45.0 & 64.1 & 49.1 & 29.5 & 47.4 & 57.8  \\
        ASAG-D (Ours)
        & ResNet-50 & 1 & 36 & 182 & 45.8 & 64.1 & 49.4 & 27.3 & 49.6 & 61.0  \\
        ASAG-A (Ours)
        & ResNet-50 & 1 & 36 & 139 & 46.3 & 65.1 & 50.3 & 29.9 & 49.2 & 59.6  \\
        
        \hline
        DETR~\cite{detr}
        & ResNet-101-DC5 & 6 & 500 & 253 & 44.9 & 64.7 & 47.7 & 23.7 & 49.5 & 62.3 \\
        SMCA~\cite{smca_detr}
        & ResNet-101 & 6 & 50 & 218 & 44.4 & 65.2 & 48.0 & 24.3 & 48.5 & 61.0  \\
        Sparse RCNN~\cite{sun2021sparse}
        & ResNet-101 & 6 & 36 & 250 & 46.4 & 64.6 & 49.5 & 28.3 & 48.3 & 61.6 \\
        Dynamic Sparse RCNN~\cite{hong2022dynamic}
        & ResNet-101 & 6 & 36 & -  & 47.8 & 67.0 & 52.0 & 31.0 & 51.1 & 62.2  \\
        Conditional DETR~\cite{conditional_detr}
        & ResNet-101-DC5 & 6 & 108 & 262 & 45.9 & 66.8 & 49.5 & 27.2 & 50.3 & 63.3 \\
        DAB-DETR~\cite{dab_detr}
        & ResNet-101-DC5 & 6 & 50 & 282 & 45.8 & 65.9 & 49.3 & 27.0 & 49.8 & 63.8   \\
        DN-DETR~\cite{dn_detr}
        & ResNet-101-DC5 & 6 & 50 & 282 & 47.3 & 67.5 & 50.8 & 28.6 & 51.5 & 65.0  \\
        AdaMixer~\cite{adamixer}
        & ResNet-101 & 6 & 36 & 201 & 48.0 & 67.0 & 52.4 & 30.0 & 51.2 & 63.7 \\
        REGO-Deformable DETR~\cite{REGO}
        & ResNet-101 & 12 & 50 & 257 & 48.5 & 67.0 & 52.4 & 29.5 & 52.0 & 64.4  \\
        SAP-DETR-DC5~\cite{SAP-detr}
        & ResNet-101-DC5 & 6 & 50 & 266 & 46.9 & 66.7 & 50.5 & 27.9 & 51.3 & 64.3  \\
        Efficient DETR~\cite{efficient_detr}
        & ResNet-101 & 1 & 36 & 289 & 45.7 & 64.1 & 49.5 & 28.2 & 49.1 & 60.2 \\
        Cascade Featurized QRCNN~\cite{zhang2022featurized}
        & ResNet-101 & 2 & 36 & 215 & 45.8 & 64.4 & 49.9 & 30.1 & 48.5 & 60.1  \\
        \hline
        ASAG-A (Ours)
        & ResNet-101 & 1 & 36 & 206 & 47.5 & 66.1 & 51.2 & 30.4 & 50.6 & 62.6 \\
        \hline
        \end{tabular}}
    \caption{Performance of different query-based detectors on COCO \texttt{minival} set with a 3$\times$ training schedule and single scale testing.
    }
    \label{tab:sota}
\end{table*}

\section{More Comparison with Other Well-Known Detectors}

In this work, we aim to narrow the performance gap between one- and six-decoder-layer detectors and retain the fast speed by Adaptive Sparse Anchor Generation. Thus the performance of our models is highly related to baselines. However, ASAGs with only one decoder layer and fewer FLOPs still provide encouraging performance compared to well-known detectors, as shown in \Cref{tab:sota}. Note that some SOTA methods propose some advanced training techniques rather than novel decoder structures and these techniques can also boost the performance of ASAG, such as denoising training~\cite{dn_detr, zhang2022dino}, more positives~\cite{group_detr, h_detr, co_detr, nms_strikes_back}, knowledge distillation~\cite{teach_detr, detrdistill, d3etr}. In \Cref{tab:dn}, we equip ASAG-A with 200 noised queries following DN-DETR~\cite{dn_detr}. The results show that Denoising Training can also benefit our methods.

\section{More Visualization}

\begin{figure*}
  \centering
  \includegraphics[width=0.87\linewidth]{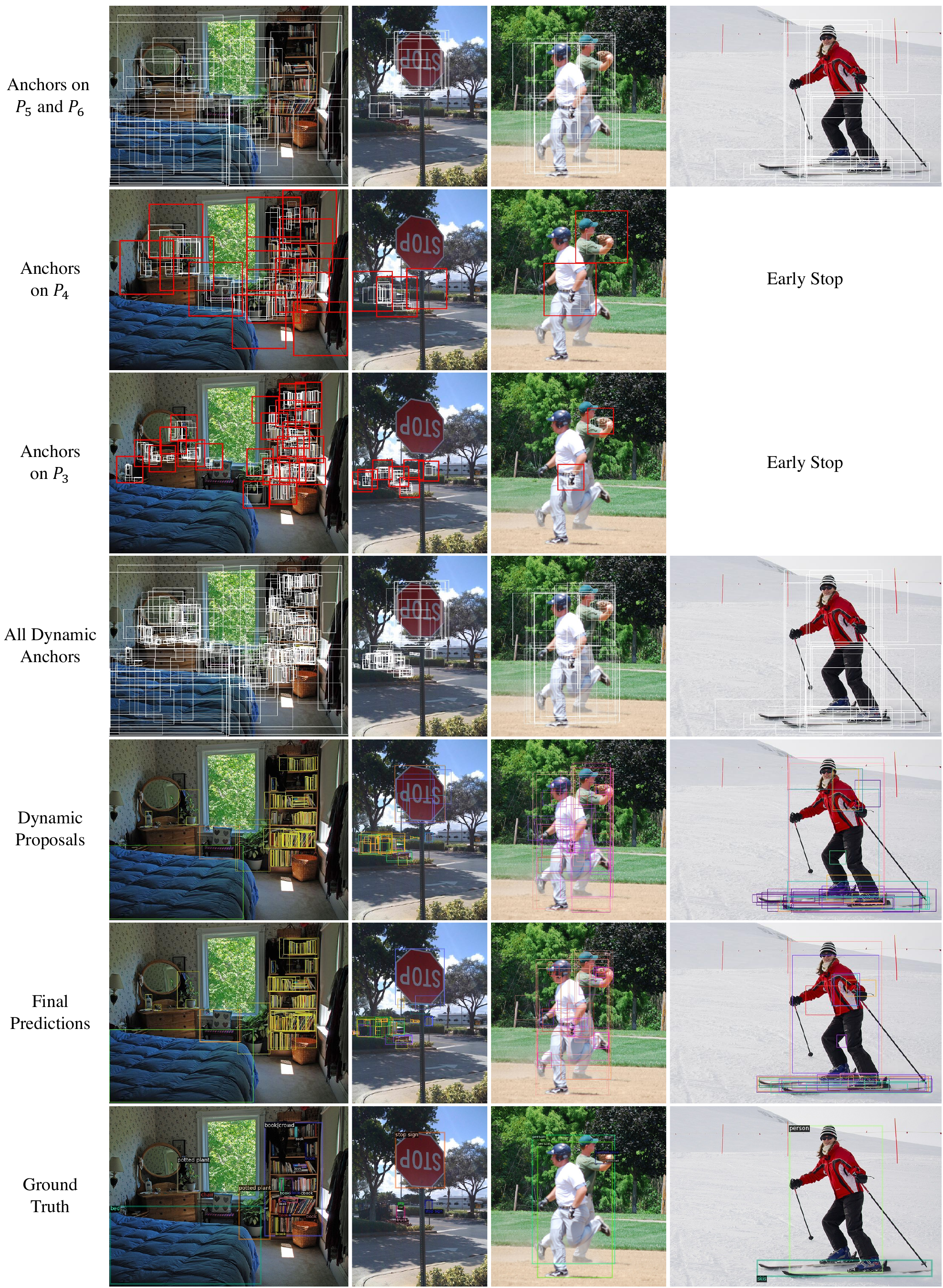}
  \caption{More visualization of bounding boxes in our pipeline. All boxes without selection are drawn in the pictures. Patches and anchors are drawn in red and white, respectively. Different colors for dynamic proposals, final predictions, and ground truth are used to separate different classes in each image.}
  \label{fig:Adaptive_Probing}
\end{figure*}

In \Cref{fig:Adaptive_Probing}, we visualize all the bounding boxes appearing through the pipeline of ASAG-A. The anchors precisely cover the foreground objects and Adaptive Probing sparsely explores large feature maps. The number of patches and the location of patches vary according to different images. In particular, the last image does not use Adaptive Probing by the early-stop mechanism since there is no small object in the image. With precise anchors, the final predictions are as close as ground truth. For the first image, we can even predict more fine-grained bounding boxes for books on the shelf than ground truth.

\begin{figure*}
  \centering
  \includegraphics[width=0.99\linewidth]{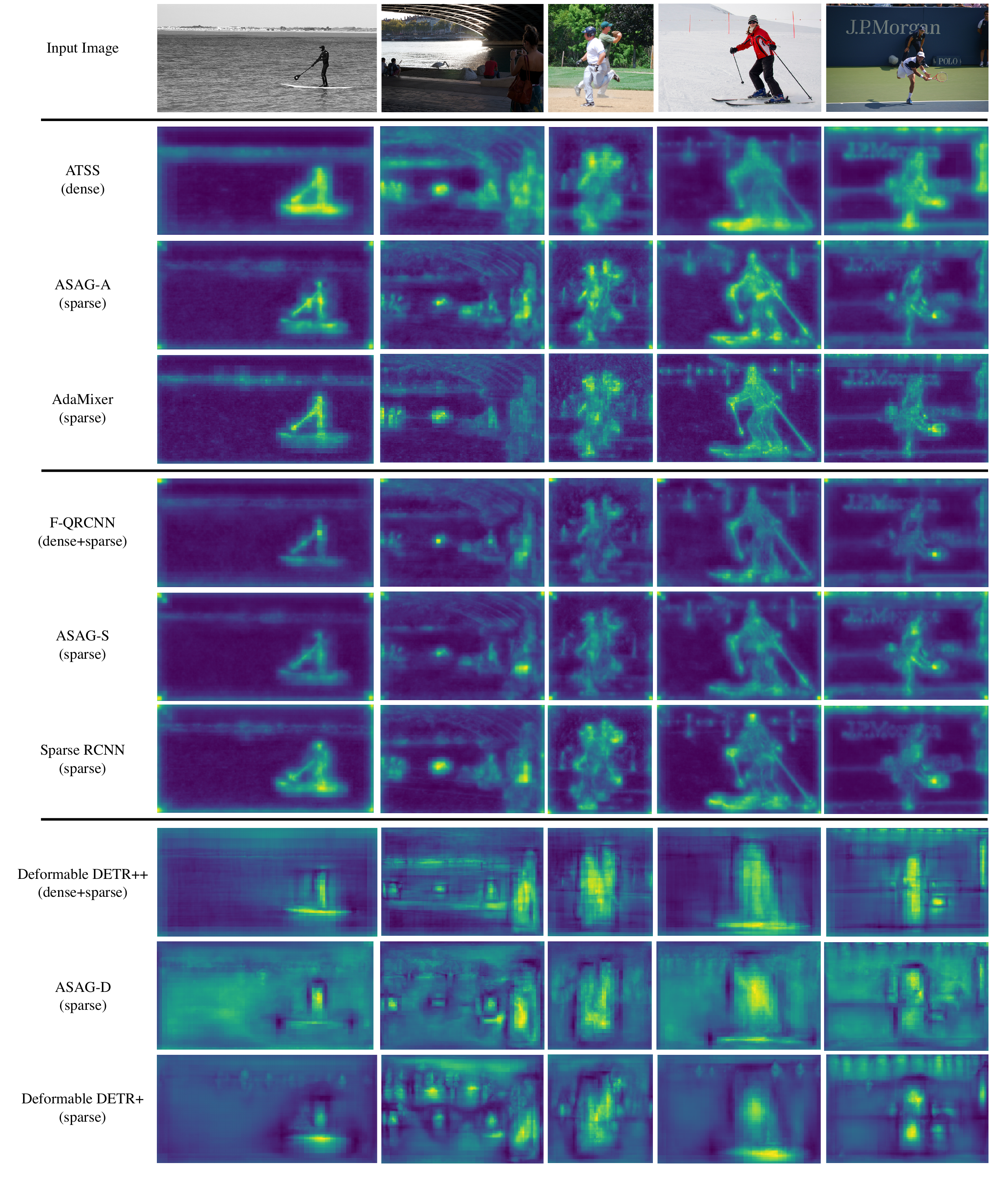}
  \caption{More visualization of feature maps. Feature maps of ASAGs with sparse initialization are more similar to six-decoder-layer sparse detectors, which highlight the discriminative parts of foreground objects.}
  \label{fig:more_feature_map}
\end{figure*}

In \Cref{fig:more_feature_map}, we compare feature maps of our models with corresponding six-decoder-layer sparse detectors and dense-initialized ones. Different from dense ones that activate the whole object uniformly, ASAGs highlight the discriminative parts of objects and pay more attention to the background, similar to six-decoder-layer sparse detectors.

\end{appendices}

\end{document}